\documentclass{article} % For LaTeX2e
\usepackage{iclr2027_conference,times}

\usepackage{amsmath,amsfonts,bm}

\def\eqref#1{equation~\ref{#1}}
\def\1{\bm{1}}

\DeclareMathAlphabet{\mathsfit}{\encodingdefault}{\sfdefault}{m}{sl}
\SetMathAlphabet{\mathsfit}{bold}{\encodingdefault}{\sfdefault}{bx}{n}

\usepackage{hyperref}
\usepackage{url}
\usepackage{xspace}

\usepackage{graphicx}
\usepackage{wrapfig}
  
\usepackage{booktabs}
\usepackage{multirow}
\usepackage{tabularx}
\usepackage{array}

\usepackage{enumitem}
\usepackage{algorithm}
\usepackage{algorithmic}

\usepackage{float}

\usepackage{xcolor}
\usepackage[most]{tcolorbox}

\definecolor{BoxBackground}{RGB}{240,240,240}
\definecolor{BoxFrame}{RGB}{0,0,0}
\definecolor{TitleBackground}{RGB}{0,0,0}
\definecolor{TitleText}{RGB}{255,255,255}

\tcbset{
academicbox/.style={
  boxsep=5pt,
  left=2pt,
  right=2pt,
  bottom=0.5pt,
  boxrule=0.5pt,
  colback=BoxBackground,
  colframe=BoxFrame,
  colbacktitle=TitleBackground,
  coltitle=TitleText,
  enhanced,
  attach boxed title to top left={
    yshift=-0.1in,
    xshift=0.1in
  },
  boxed title style={
    boxrule=0pt,
    colframe=white
  },
  title={#1},
}
}

\usepackage{xcolor}
\usepackage{listings}

\definecolor{jsonblue}{RGB}{36, 99, 160}
\definecolor{jsonorange}{RGB}{196, 94, 0}
\definecolor{jsonbackground}{RGB}{247, 247, 247}
\definecolor{jsonframe}{RGB}{210, 210, 210}

\lstdefinelanguage{json}{
  basicstyle=\ttfamily\footnotesize,
  numbers=none,
  showstringspaces=false,
  breaklines=true,
  breakatwhitespace=false,
  columns=fullflexible,
  keepspaces=true,
  frame=single,
  rulecolor=\color{jsonframe},
  backgroundcolor=\color{jsonbackground},
  stringstyle=\color{jsonblue},
  keywordstyle=\color{jsonorange},
  commentstyle=\color{gray},
  tabsize=2,
  xleftmargin=4pt,
  xrightmargin=4pt,
  framexleftmargin=4pt,
  framexrightmargin=4pt
}

\newtcolorbox{AcademicBox}[1][]{academicbox=#1}

\usepackage{graphicx}
\def\onedot{.\xspace}
\def\ie{\emph{i.e}\onedot} 
\def\eg{\emph{e.g}\onedot} 

\definecolor{cvprblue}{rgb}{0.21,0.49,0.74}
\hypersetup{colorlinks,urlcolor=black,citecolor=cvprblue}
\title{WorldPlay2: Extending Real-Time Interactive World Models in Control and Horizon}

\author{%
    \parbox[t]{\dimexpr\textwidth-2\tabcolsep\relax}{%
      \centering
      \textbf{%
        Haiyu Zhang\thanks{Equal contribution.}
        \quad Wenqiang Sun\textsuperscript{*}
        \quad Tengfei Wang\thanks{Corresponding authors.}
        \quad Junta Wu
        \quad Jun Zhang
      }\\[0.6em]
      \textbf{%
        Yunhong Wang
        \qquad Yu Qiao
        \qquad Chunchao Guo\textsuperscript{$\dagger$}
      }%
      \\[0.8em]
      \normalfont\small
        Project page:
    \href{https://worldplay2.github.io/}{%
      \textcolor{red}{\nolinkurl{https://worldplay2.github.io/}}}
    }%
}

\iclrfinalcopy % Uncomment for camera-ready version, but NOT for submission.
\begin{document}

\maketitle
\fancyhead{}                       % 清除页眉文字
\renewcommand{\headrulewidth}{0pt}  % 去掉页眉横线

\begin{figure}[H]
  \vspace{-6mm}
  \centering
  \includegraphics[width=\linewidth]{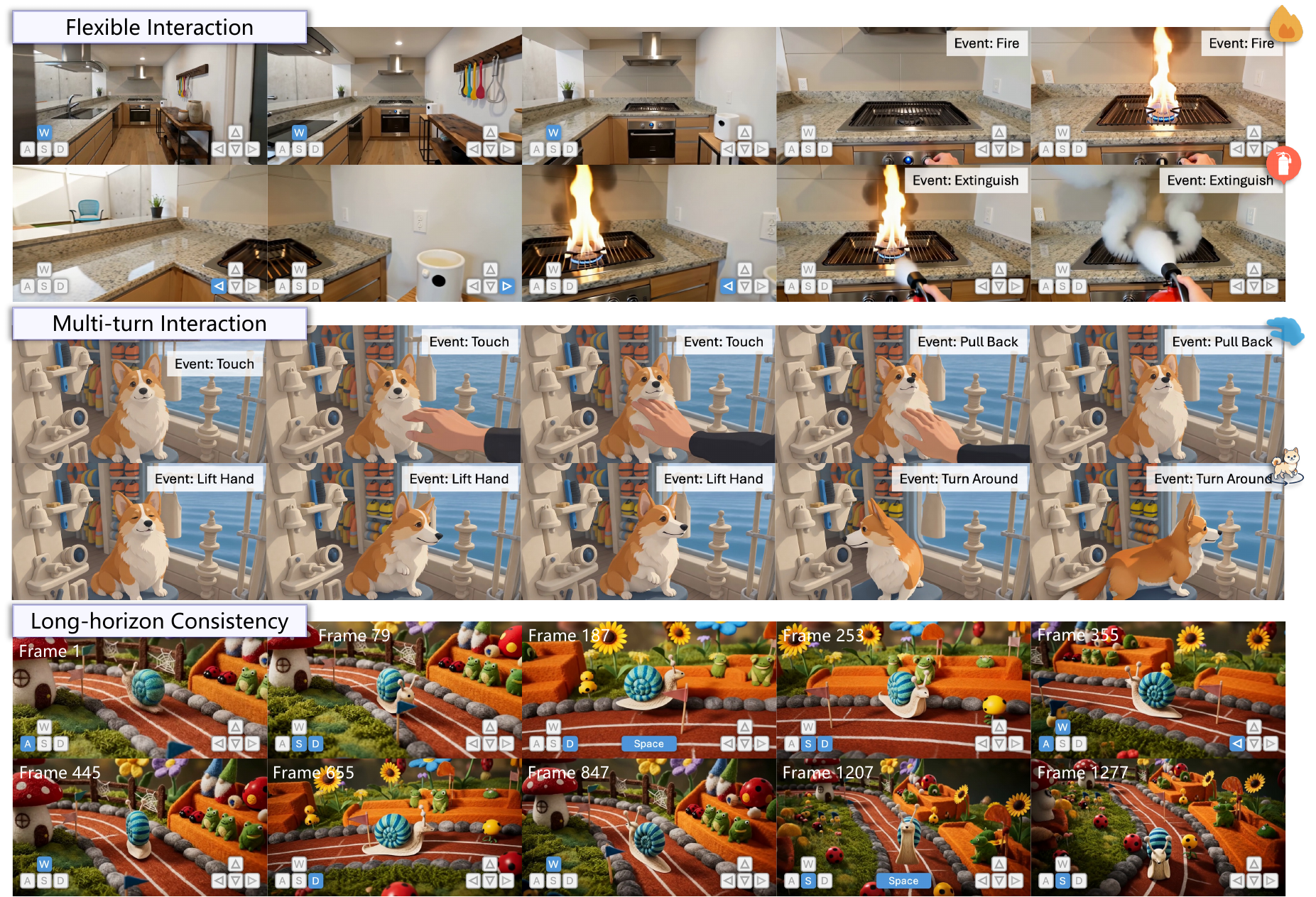}
  \vspace{-6mm}
  \caption{\textbf{WorldPlay2 is a real-time interactive world model enabling versatile controls with long-horizon consistency.} 
    \textbf{Top:} It supports flexible interactions, including navigation controls and complex semantic events. \textbf{Middle:} It executes multi-turn interactions to achieve coherent storytelling. \textbf{Bottom:} It maintains long-horizon consistency under complex navigation controls. }
  \label{fig:teaser}
\end{figure}
  
\begin{abstract}
Interactive world models require responding in real time to versatile controls and maintaining long-horizon consistency.
However, modeling heterogeneous controls remains difficult, while explosive contexts and unstable distillation impede achieving both long-horizon consistency and real-time responsiveness. In this paper, we present WorldPlay2, an interactive world model that couples a factorized hybrid control interface with a co-design of compressed memory and stable distillation. 1) Our factorized hybrid control interface integrates frame-aligned action control with structured semantic control that explicitly disentangles scene appearance, character identity, and dynamic semantic events, thereby facilitating effective control learning. 2) To achieve efficient long-horizon modeling, we compress historical contexts into compact memory tokens shared by the autoregressive student and the bidirectional teacher. This design enables clip-wise, memory-conditioned score evaluation instead of jointly processing an entire long rollout, substantially reducing distillation overhead.  3) We further propose Stable Forcing, which initializes the autoregressive student via a few-step strategy and leverages full-rollout replay to preserve the quality of long-horizon rollouts, ensuring robust and stable distillation. Extensive experiments demonstrate the strong generalizability of our model and its superior performance compared to existing methods.
\end{abstract}

\section{Introduction}

Interactive world models~\citep{parker2025genie, sun2025worldplay, he2025matrix, team2026advancing, happyoyster, hong2025relic, xu2026wonder, jiang2026abot, zhu2026sana, wang2026worldcompass} are beginning to transform video generators~\citep{wan2025wan, wu2025hunyuanvideo, veo} from passive content-creation systems into interactive environments that evolve in response to user input. Such models have the potential to serve as general-purpose simulators~\citep{agarwal2026cosmos, brooks2024video}, allowing users to explore, interact with, and reshape the generated environments while providing scalable data generators and policy evaluators for embodied agents~\citep{wiedemer2025video, team2025evaluating}. Realizing these applications requires world models to support versatile controls, preserve coherent world states over long horizons, and operate in real time.

Controllability represents a primary challenge for world models. Early efforts~\citep{sun2025worldplay, he2025matrix, team2026advancing} focused on navigation-oriented controls, lacking richer mechanisms for interacting with the environment. Although recent works~\citep{team2026alayaworld, gao2026infinite} attempt to accommodate increasingly diverse interactive events, effectively integrating these heterogeneous inputs remains challenging because these control signals inherently operate across disparate semantic granularities and temporal horizons. Specifically, camera motion and character locomotion demand precise, frame-aligned control, whereas complex interactive events are more naturally expressed via high-level semantic commands spanning longer temporal horizons. Moreover, control signals must be disentangled from content factors such as scene appearance and character identity. Otherwise, the model may conflate visual appearance, spatial movement, and occurring events, leading to ambiguous supervision and unreliable responses.

The second challenge lies in co-designing memory and distillation. Existing methods~\citep{team2026advancing, hong2025relic, xu2026wonder, zhu2026sana} often rely on full-context bidirectional models as teachers. However, computational costs scale quadratically with video length, which not only hinders long-horizon modeling but, more crucially, makes score evaluation during distillation prohibitively expensive. Moreover, preserving long-horizon consistency during distillation presents additional challenges. This process typically requires the student model to autoregressively generate long video sequences~\citep{hong2025relic, sun2025worldplay, xu2026wonder, team2026advancing}, but due to error accumulation and few-step sampling, the student's generation distribution diverges significantly from that of the teacher, thereby rendering distillation unstable and severely degrading generation quality. Therefore, world models require a co-design in which memory is sufficiently compact to enable efficient long-horizon modeling and distillation is sufficiently stable to preserve long-horizon consistency.

In this paper, we introduce WorldPlay2, a real-time interactive world model that combines factorized hybrid control interface, distillation-oriented compressed memory, and stable long-horizon distillation. Specifically, 
we propose a factorized hybrid control interface that explicitly disentangles low-level movements, high-level semantic interactions, and visual content factors. Low-level movements, \eg, camera motion and third-person character locomotion, are precisely modulated via frame-aligned action control. Concurrently, to govern high-level semantic interactions and content factors, we design a structured semantic control that partitions signals into three decoupled fields, \ie, scene, character, and event, where each field governs a distinct concept within the world.
By disentangling these signals, the model learns reusable combinations across the heterogeneous controls while retaining the precise responsiveness required for world models.

Then, we co-design the memory and distillation. For the memory mechanism, we compress the generated history into compact memory tokens, substantially reducing the computational cost for long-horizon modeling. Crucially, this design bypasses score evaluations across the full-resolution rollouts during distillation. By partitioning long-horizon rollouts into local temporal clips conditioned on compact memory tokens, we can compute scores independently per clip, ensuring scalable and computationally tractable distillation.
For distillation, we propose Stable Forcing, a stable framework tailored for long-horizon distillation. We first warm-start the autoregressive student via a few-step initialization scheme inspired by PDD~\citep{shaul2026parallel}, yielding a well-behaved few-step student. This initialization ensures that the student's rollout distribution closely aligns with that of the teacher over long horizons, thereby stabilizing subsequent distribution distillation. Building on this starting point, we perform distribution-matching distillation to enhance long-horizon consistency and mitigate exposure bias. In this stage, we introduce full-rollout replay to decouple long-horizon rollouts from gradient backpropagation. During the forward rollout phase, each chunk undergoes full few-step sampling to maintain fidelity and bolster stability. Meanwhile, a random intermediate step is recorded and replayed with gradients during the backward pass. Combining our initialization with full-rollout replay preserves rollout quality over long horizons, thereby achieving stable and robust distillation. 

Taken together, our model demonstrates remarkable generalization across different scenes and characters. As shown in Fig.~\ref{fig:teaser}, it not only supports versatile, multi-turn interactive controls, but also preserves geometric consistency over long horizons. Moreover, extensive quantitative and qualitative experiments validate the effectiveness of our methods, demonstrating superior performance compared to existing methods.

\section{Related Work}

\textbf{Interactive World Models.} Interactive world models generate future visual frames conditioned on previous observations and current actions, enabling users or embodied agents to interact with the environment. Recent world models have substantially expanded environmental diversity~\citep{zhang2025matrix, he2025matrix, li2025hunyuan, team2026dreamx, mao2025yume, jiang2026abot, hunyuanworld_2025, hyworld2_2026}, long-horizon consistency~\citep{sun2025worldplay, hong2025relic, xu2026wonder, team2026advancing, wang2026matrix, team2026inspatio}, and the range of supported controls~\citep{parker2025genie, gao2026infinite, team2026alayaworld, mao2026yume1, happyoyster, tang2025hunyuan}. WorldPlay~\citep{sun2025worldplay}, Lingbot-World~\citep{team2026advancing}, and Wonder~\citep{xu2026wonder} utilize camera poses, discrete keyboard inputs, or pixel-space coordinate field as control signals to govern viewpoint transformation and character movement. Lingbot-World-V2~\citep{gao2026infinite} and AlayaWorld~\citep{team2026alayaworld} introduce language-driven events to further enable richer interactive controls. However, these control signals are inherently heterogeneous, making unified and effective representation particularly challenging. Moreover, existing methods model long-horizon consistency via retrieval~\citep{yu2025context, xiao2026worldmem}, sparse attention~\citep{xu2026wonder}, or explicit 3D representations~\citep{team2026inspatio, team2026alayaworld}, while treating distillation as an isolated module. In contrast, our method co-designs the memory mechanism and distillation, improving both training efficiency and distillation stability.

\textbf{Distillation.} Distillation accelerates diffusion models by reducing the number of function evaluations. One representative line of studies aggregates multi-step instantaneous velocity into single-step average velocity. For instance, MeanFlow~\citep{geng2026mean} derives the relationship between instantaneous and average velocities, whereas rCM~\citep{zheng2026large}, AnyFlow~\citep{gu2026anyflow}, and TiM~\citep{wang2026transition} implement a parallelism-compatible JVP kernel or differential derivation to scale this approach to large-scale models. PiFlow~\citep{chen2026pi} and PDD~\citep{shaul2026parallel} further optimize trajectory learning to estimate average velocities more efficiently, achieving strong performance in bidirectional model distillation. Another major paradigm performs distribution matching distillation~\citep{yin2024one, yin2024improved, yin2025slow, zheng2026causal, zhu2026causal, huang2026self}, aligning the generation distribution of a few-step student with that of a multi-step teacher. This paradigm is widely adopted for interactive world models because it accelerates sampling, mitigates exposure bias, and inherits desirable properties from the teacher, such as long-horizon consistency. However, when the student performs few-step long-horizon rollouts, its generation distribution diverges significantly from that of the teacher, making distillation highly unstable. 

\begin{figure*}[t]
  \centering
  \includegraphics[width=\textwidth]{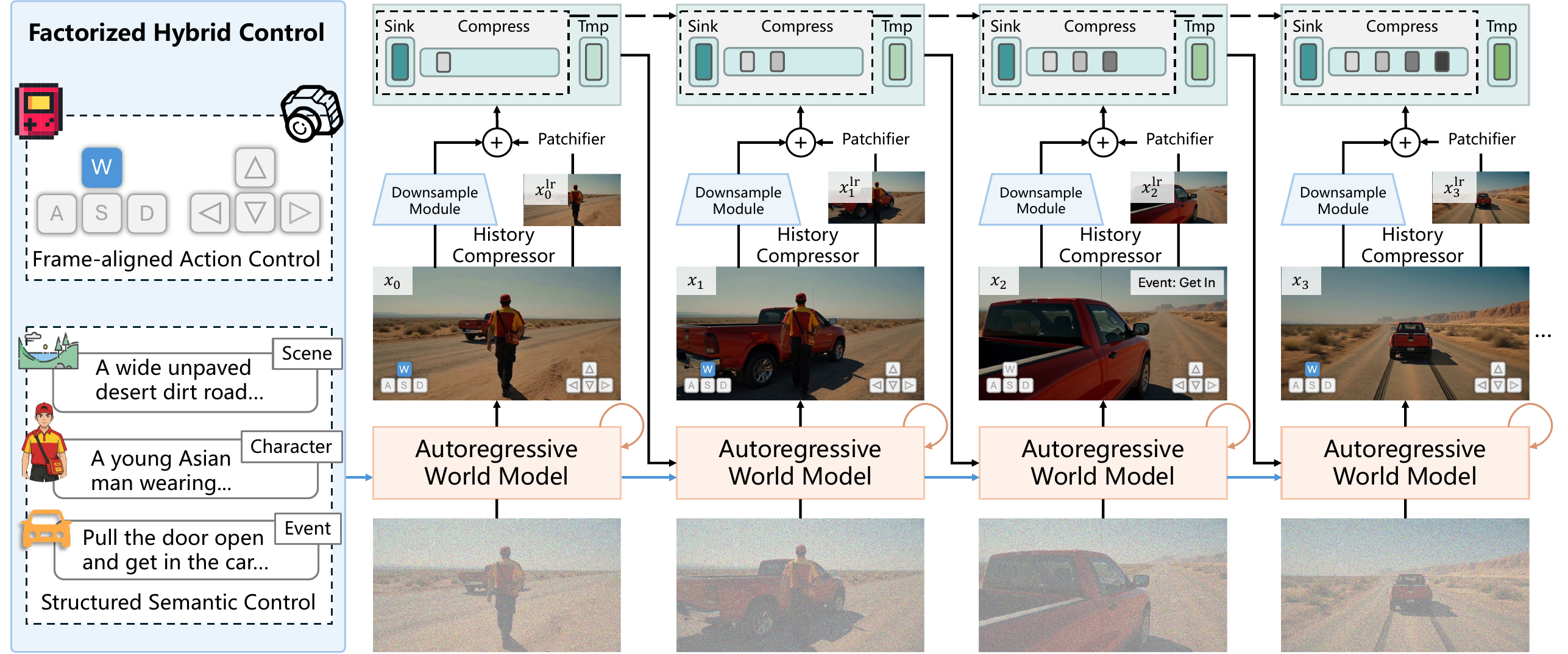}
  \vspace{-7mm}
  \caption{\textbf{Overview of WorldPlay2.} We employ factorized hybrid control interface to decouple heterogeneous inputs into frame-aligned action control and structured semantic control, ensuring accurate interactive responses. Concurrently, a history compressor abstracts past contexts into compact memory tokens to achieve efficient long-horizon inference.}
  \label{fig:pipeline}
  \vspace{-3mm}
\end{figure*}

\section{Method}

Our goal is to construct a real-time interactive world model $N_{\theta}(x_{t}|x_{<t}, A_{\le t})$ parameterized by $\theta$ that supports versatile controls while maintaining long-horizon consistency. The model generates next chunk $x_{t}\in \mathbb{R}^{T\times H \times W}$ based on past observations $x_{<t}=\{x_{t-1}, ...,x_{0}\}$, controls $A_{<t}=\{A_{t-1},...,A_{0}\}$, and current control signal $A_{t}$. We first introduce our factorized hybrid control interface in Sec.~\ref{method_control}, which disentangles heterogeneous inputs to support diverse interactions. In Sec.~\ref{method_memory}, we present our distillation-oriented compressed memory mechanism, enabling efficient long-horizon modeling and  teacher supervision. Finally, we detail Stable Forcing in Sec.~\ref{method_distillation}, a stable long-horizon distillation framework that distills a many-step autoregressive model into a few-step model while preserving long-horizon consistency. Fig.~\ref{fig:pipeline} illustrates the overview of our model.

\subsection{Factorized Hybrid Control Interface}
\label{method_control}

Interactive world models require versatile responsiveness to heterogeneous controls, which often exhibit varying levels of abstraction. Camera motion and character locomotion demand precise, frame-aligned signals, whereas complex interactions and content factors are more naturally expressed via high-level semantic instructions. Therefore, we propose a factorized hybrid control interface that disentangles low-level movements, high-level semantic interactions, and visual content factors. Specifically, each control signal is represented as $A_t = \{a_t, d_t\}$, where $a_t$ denotes frame-aligned action control and $d_t$ denotes structured semantic control.

\textbf{Frame-aligned Action Control.} Our low-level movements $a_{t}$ consist of continuous camera pitch and yaw angles, discrete longitudinal and lateral movements, the camera perspective, and a special action (\ie, jumping).
We separately embed the continuous and discrete components and combine them into a unified action representation,
\begin{equation}
e_t =
E(\text{continuous})+ \mathbf{e}[\text{discrete}],
\label{eq:action_embedding}
\end{equation}
where $E$ is an MLP for continuous signals and $\mathbf{e}$ denotes the learnable embeddings for the discrete signals. 
The resulting action representation is aligned with the corresponding visual tokens and injected before the feed-forward network (FFN) in each Transformer block,
\begin{equation}
\tilde{h}_t =
h_t + F\left([h_t\oplus e_t]\right),
\label{eq:action_injection}
\end{equation}
where $F$ is an auxiliary MLP, $h_t$ denotes the hidden states, and $\oplus$ means channel concatenation.

\textbf{Structured Semantic Control.} High-level semantic control involves diverse interactions and visual content factors that often span longer temporal horizons, making it challenging to represent with low-dimensional vectors. Therefore, we utilize structured caption $d_t$ as the semantic control signals. Specifically, $d_t$ encapsulates visual content factors (\ie, scene and character identity) as well as dynamic semantic events,
\begin{equation}
d_t =
(
d_{\text{scene}},
d_{\text{character}},
d_{\text{event}}
).
\end{equation}
Here, the scene field $d_{\text{scene}}$ describes the environmental content, including spatial layout, objects, illumination, and visual style. $d_{\text{character}}$ specifies the persistent identity and appearance of the controlled entity. $d_{\text{event}}$ describes the semantic change associated with the current event, such as object manipulation, environmental changes, and object appearance.

\subsection{Distillation-Oriented Compressed Memory}
\label{method_memory}
Long-horizon world modeling requires access to historical information beyond a limited local context. A straightforward approach is to retain the entire full-resolution contexts~\citep{team2026advancing, hong2025relic}. However, this causes context length to scale linearly during autoregressive rollout, rapidly increasing inference latency and complicating long-horizon modeling. Moreover, this computational burden is further amplified during distillation, where the full-context bidirectional teacher is used to evaluate long student rollouts to provide supervision. To this end, we design a compressed memory mechanism inspired by~\cite{zhang2025tinyhistory}, shared across long-horizon modeling and distillation. This enables efficient context conditioning while keeping teacher evaluation during distillation computationally tractable.

Instead of conditioning the model directly on the full-resolution contexts, we employ a learnable history compressor $\mathcal{C}_{\phi}$ to encode the history context into a compact sequence of memory tokens:
\begin{equation}
    m_{<t}=\left[
   x_{\text{sink}};
  x_{\text{cmp}}=\mathcal{C}_{\phi}(x_{<t}, x_{<t}^{\text{lr}});
  x_{\text{tmp}}
  \right],
\end{equation}
\begin{wrapfigure}{r}{0.38\textwidth}
  \centering
  \vspace{-10pt}
  \includegraphics[width=\linewidth]{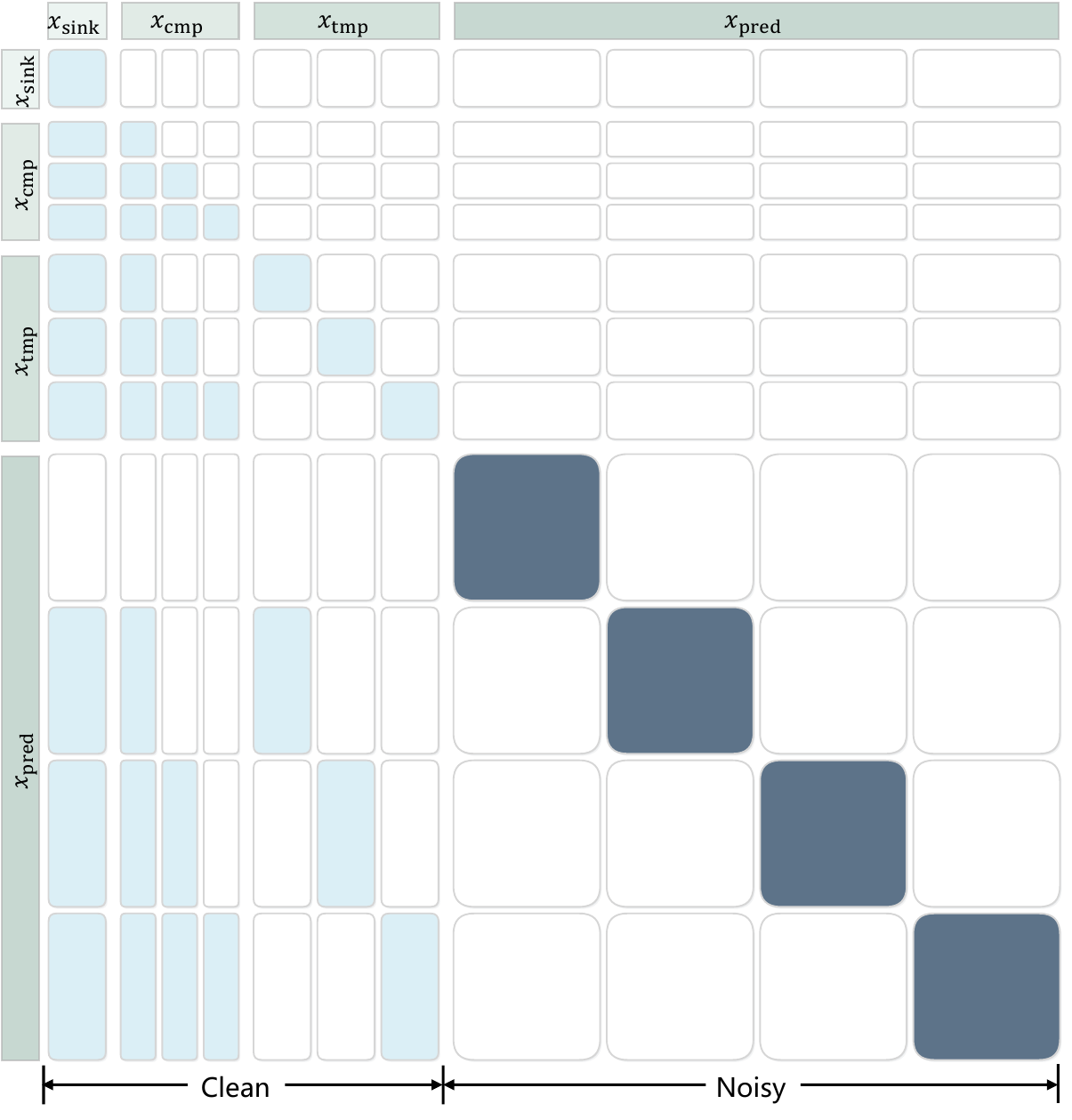}
  \vspace{-24pt}
  \caption{Causal attention mask for compressed memory training.}
  \label{fig:teacher_forcing_mask}
  \vspace{-16pt}
\end{wrapfigure}
where $[;;]$ denotes sequence concatenation, $x_{\text{sink}}$ denotes sink tokens providing a stable reference, $x_{\text{tmp}}$ represents adjacent temporal tokens that enforce temporal consistency, and $x_{<t}^{\text{lr}}\in \mathbb{R}^{\frac{T}{l}\times \frac{H}{s}\times \frac{W}{s}}$ denotes the low-resolution, low-frame-rate video latent encoded by the VAE, with $l=2$ and $s=4$ representing the temporal and spatial downsampling factors, respectively. For the history compressor $\mathcal{C}_{\phi}$, we adopt a dual-branch design~\citep{zhang2025tinyhistory}. 
Specifically, the coarse branch processes $x_{<t}^{\text{lr}}$ through the DiT's patchifier to produce coarse features, while the fine branch passes $x_{<t}$ through a downsample module to yield residual fine features. This dual-branch representation provides both high-level semantic information and fine-grained visual details for subsequent generation. Compared to full-resolution contexts, our memory compression mechanism reduces the sequence length by a factor of approximately $l \cdot s^2=32$, significantly reducing computational overhead.

Given a long training video, we partition it into a compressed historical context and a target clip $x_{[t:t+L]}$. For the causal autoregressive student $N_{\theta}$, we employ teacher forcing with a causal attention mask (illustrated in Fig.~\ref{fig:teacher_forcing_mask}) to maintain temporal causality and follow the flow matching objective~\citep{lipman2022flow},
\begin{equation}
\mathcal{L}_{\text{student}} = \mathbb{E}_{x, \boldsymbol{\epsilon}, \sigma} \left\| N_{\theta} \left( x_{[t:t+L]}^{\sigma}, \sigma, m_{< t+L}, A_{\le t+L} \right) - (\boldsymbol{\epsilon} - x_{[t:t+L]}) \right\|^2,
\end{equation}
where $\boldsymbol{\epsilon} \sim \mathcal{N}(0, \mathbf{I})$ denotes Gaussian noise and $\sigma$ represents the diffusion noise level. Concurrently, the bidirectional teacher model is trained without the causal attention mask.
By applying our proposed memory mechanism to both student and teacher models, we can efficiently model long-horizon consistency. Moreover, it enables long student rollouts to be partitioned into smaller clips for individual teacher evaluation, significantly reducing computational overhead during distillation.

\subsection{Stable Forcing}
\label{method_distillation}
\begin{figure*}[t]
  \centering
  \includegraphics[width=\textwidth]{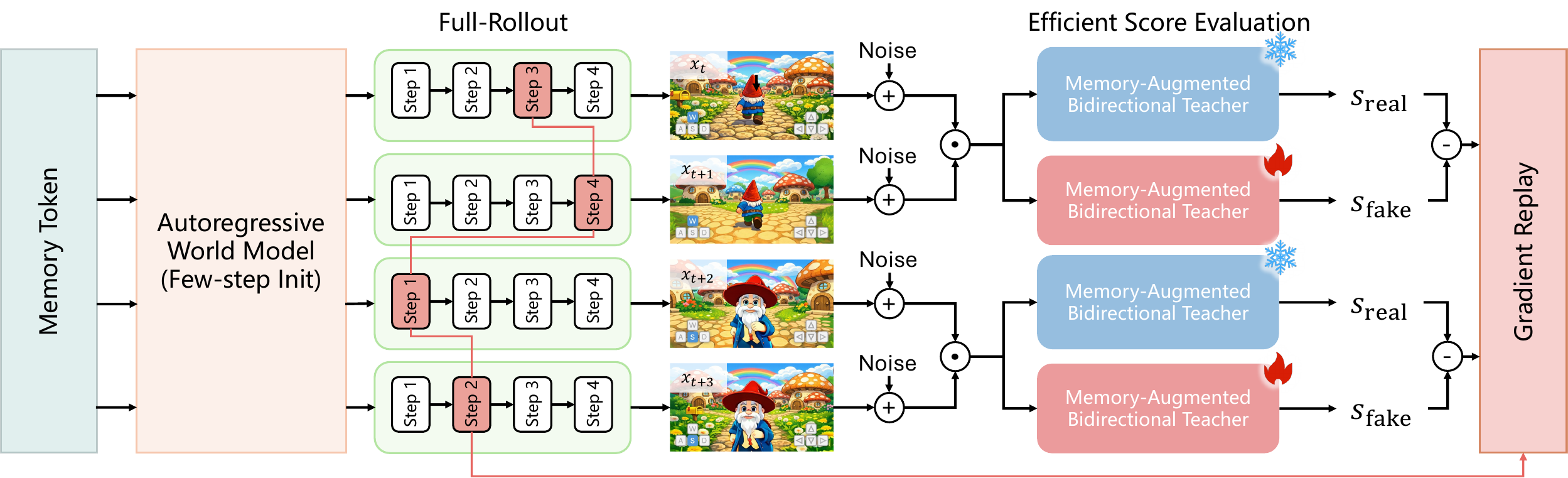}
  \vspace{-7mm}
  \caption{\textbf{Overview of Stable Forcing.} It integrates few-step initialization, full-rollout replay, and efficient score evaluation to achieve efficient and stable long-horizon distillation.}
  \label{fig:stable_forcing}
  % \vspace{-4mm}
\end{figure*}

Self Forcing~\citep{huang2026self} has emerged as an effective approach for distilling autoregressive video diffusion models, as it simultaneously reduces sampling steps and mitigates error accumulation. However, extending it to long-horizon rollout introduces new challenges. Performing long-horizon student rollouts with few sampling steps causes errors to compound rapidly across chunks, driving the student's generation distribution away from the teacher's and leading to unstable training. Additionally, evaluating scores over long rollouts using full-context teacher models demands substantial computational resources and GPU memory. To address these challenges, we introduce Stable Forcing as shown in Fig.~\ref{fig:stable_forcing}, a long-horizon distillation framework designed to achieve stable training via few-step initialization, full-rollout replay, and efficient score evaluation.

\textbf{Few-step Initialization.} Stable long-horizon distillation requires the student to produce meaningful rollouts under few-step sampling. To establish a reliable few-step initialization, we extend PDD~\citep{shaul2026parallel} to our memory-augmented autoregressive student model. Specifically, PDD discretizes the diffusion noise schedule into $K$ blocks, where each block $i$ contains $C$ sub-intervals $\{\sigma_{0}^{i}, \dots, \sigma_{C-1}^{i}\}$. A parallel decoder then predicts the mean velocities $u_{0},...,u_{C-1}$ across adjacent intervals within a block in a single forward pass. The training objective is,
\begin{equation}
    \mathcal{L}_{\text{PDD}}=\mathbb{E}_{k\in[0,C-1]} \left\| u_{k}  - \text{sg}(N_{\theta}(x_{t}^{\sigma_{k}^i},\sigma_{k}^i,m_{<t},A_{\le t})) \right\|^2,
\end{equation}
where $x_{t}^{\sigma_{k}^i}$ is computed via parallel decoding sampling and $\text{sg}(\cdot)$ denotes the stop-gradient operator.
By predicting multiple consecutive denoising intervals in parallel, it reduces the number of network evaluations and provides a reliable few-step initialization to stabilize subsequent long-horizon distribution matching distillation.

\textbf{Full-rollout Replay.} To further enhance the stability of distribution matching distillation, we decouple the rollout phase from gradient backpropagation. Specifically, during the rollout phase, each chunk performs full few-step sampling, and only its final prediction is incorporated into subsequent chunk generation and score evaluation. Simultaneously, we cache a randomly selected intermediate denoising timestep for each chunk and replay it with gradients after computing the score. This strategy not only improves rollout quality but also ensures supervision across denoising timesteps.

\textbf{Efficient Score Evaluation.} After the student model generates a long rollout $x_{[0:BL]}$, we perform an efficient score evaluation to obtain the distribution-matching signal. Specifically, the rollout is partitioned into $B$ clips. For each clip $x_{[iL:(i+1)L]}$, the preceding chunks are encoded into compact memory tokens $m_{<iL}$ and the real and fake scores are evaluated using the teacher model $v$ as follows,
\begin{equation}
    s_{\text{fake/real}}=v(x_{[iL:(i+1)L]}^{\sigma}, \sigma, m_{<iL}, A_{\le (i+1)L}).
\end{equation}
In this manner, we preserve long-horizon supervision while reducing the sequence length processed by the score model, thereby achieving more efficient score evaluation.

\vspace{-2mm}
\section{Experiments}

\textbf{Dataset.} Our training corpus comprises two distinct subsets: a spatial navigation dataset and an interactive event dataset. Our navigation dataset aggregates various sources, including SpatialVID~\citep{wang2026spatialvid}, Sekai~\citep{li2026sekai}, ABot-World~\citep{jiang2026abot}, internal gameplay recordings, and Unreal Engine (UE) rendering sequences, totaling 700K video clips (lasting 30s to 60s). To endow the model with flexible interactive capabilities, we construct an interactive event dataset comprising 10K clips (lasting 10s to 30s). This subset covers three categories, \ie, environmental transition, object addition/removal, and complex interaction. Although the interactive event dataset is relatively small, the pretrained model inherently exhibits strong instruction-following capability. Therefore, it suffices to unlock this capability. Details are provided in the Appendix.

\begin{table*}[t]

\caption{
\textbf{Quantitative comparisons.} We benchmark our approach against recent interactive world models on both WBench and RevisitBench to systematically evaluate controllability, long-horizon consistency, and visual fidelity. \textit{Abbreviations: Avg: Average, Qua: Quality, Set: Setting, Int: Interaction, Con: Consistency, Phy: Physical.}
% \textbf{\underline{Bold and underline}} presents the 1st, \textbf{Bold} indicates the 2nd, and \underline{underline} means the 3rd.
}
\label{tab:long-term}
% \vspace{-1mm}
\centering
\setlength{\tabcolsep}{3pt} % <--- 加上这行，调节列间距（数值越小越紧凑）
\scriptsize
% \tiny
\renewcommand{\arraystretch}{1.1}
\newcommand{\gou}{\textcolor{ForestGreen}{\ding{52}}}
\newcommand{\cha}{\textcolor{Red}{\ding{55}}}
\scalebox{1.0}{
\begin{tabular}{lcccccc|cccc}
\toprule
\multicolumn{2}{c}{} &
\multicolumn{5}{c}{\textbf{WBench}} &
\multicolumn{4}{c}{\textbf{RevisitBench}} \\
\cmidrule(lr){2-7}\cmidrule(lr){8-11}
\multicolumn{1}{c}{} &
% \textbf{AVERAGE} $\uparrow$ &
% \textbf{QUALITY} $\uparrow$ &
% \textbf{SETTING} $\uparrow$ &
% \textbf{INTERACTION} $\downarrow$ &
% \textbf{CONSISTENCY} $\downarrow$ &
% \textbf{PHYSICAL} $\downarrow$ &
\textbf{Avg.} $\uparrow$ &
\textbf{Qua.} $\uparrow$ &
\textbf{Set.} $\uparrow$ &
\textbf{Int.} $\uparrow$ &
\textbf{Con.} $\uparrow$ &
\textbf{Phy.} $\uparrow$ &
\textbf{PSNR} $\uparrow$ &
\textbf{SSIM} $\uparrow$ &
\textbf{LPIPS} $\downarrow$ &
\textbf{MEt3R} $\downarrow$ \\
\midrule
   WorldPlay~\citep{sun2025worldplay}
     &  78.1 & 78.1  & 72.2  & 86.8 & 86.9
    & 66.3 &  17.05 & 0.553  & 0.416 & 0.179 \\

    AlayaWorld~\citep{team2026alayaworld}
     & 76.3  & 79.3   & 69.7  & 80.0 & 89.5
    & 63.1 & 13.61  & 0.399  & 0.569 & 0.244 \\

   Lingbot-World-V2~\citep{gao2026infinite}
    & 79.4 & 81.8  & 76.8 & 82.8 & 86.5
    & 69.1 & 14.46  & 0.469 & 0.506 & 0.301 \\

    EchoWM~\citep{zhang2026echowm}
    & 81.0 & 81.1   & 77.5 & {87.9} & 88.3 
    & 70.1 & 14.83  & 0.513 & 0.477 & 0.283  \\
    
   Alaya-Evoke-Turbo~\citep{yin2026alaya}
    & 82.0 & \textbf{81.9}  & \textbf{82.1} & 83.9 & 88.1
    & {74.0} & 16.33  & 0.501  & 0.412 & 0.236 \\

    \midrule

    % Ours-teacher
    % & \textbf{22.09} & {0.687}  & \textbf{0.219}   & {0.028}& \textbf{0.113}
    % & \textbf{19.31} & \textbf{0.599} & 0.383 & \textbf{0.209} & \textbf{0.717}  \\

    Ours (w/o Stable Forcing)
     & 79.0 & 79.1  & 77.8 & 82.4 & 87.4
    & 68.2 & 16.52 & 0.535 & 0.439 & 0.215  \\

    Ours (full)
    & \textbf{83.1} & 81.8 & 81.5 & \textbf{88.3}  & \textbf{90.0}
    & \textbf{74.0} & \textbf{19.71} &  \textbf{0.613}  & \textbf{0.318} & \textbf{0.105}  \\
\bottomrule
\end{tabular}
}
\vspace{-0mm}
% \caption{
% \textbf{Quantitative comparison results.} Our model surpasses almost all baselines in metrics encompassing visual quality, long-term consistency, and camera control accuracy. 
% }

\vspace{-4mm}
\end{table*}
\begin{figure*}[t]
  \centering
  \includegraphics[width=\textwidth]{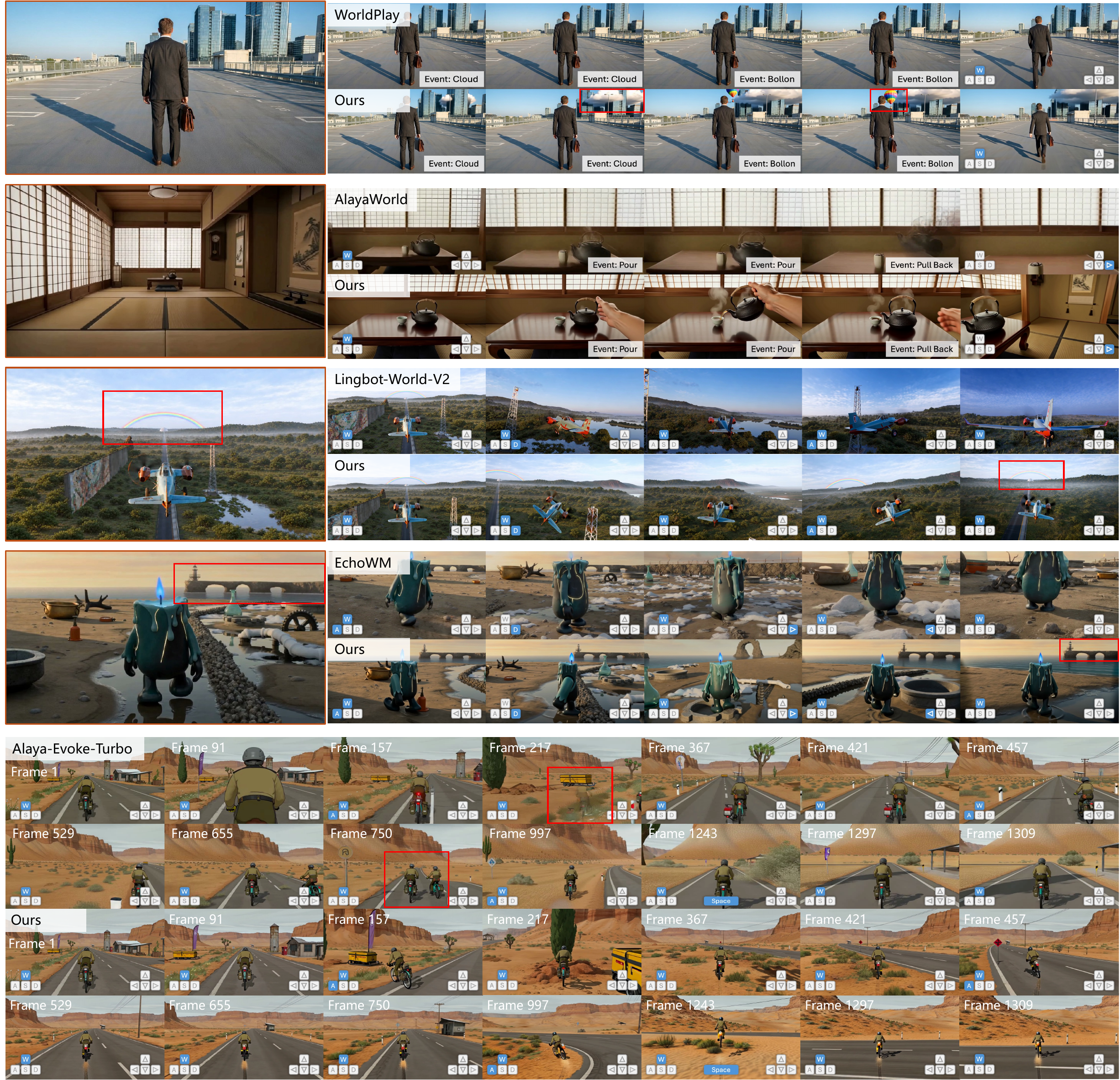}
  \vspace{-5mm}
  \caption{\textbf{Qualitative comparisons with existing methods.} WorldPlay2 supports versatile interactive events while demonstrating superior generalization across diverse characters and maintaining long-horizon geometric consistency.}
  \label{fig:main_results}
  \vspace{-2mm}
\end{figure*}

\textbf{Implementation Details.}  Our training follows a multi-stage curriculum. Specifically, the base video diffusion model is first trained for navigation controls on our spatial navigation dataset. Subsequently, utilizing the full dataset, we integrate the memory compressor following the two-stage training regime of \cite{zhang2025tinyhistory} to improve long-horizon geometric consistency. Next, the bidirectional model is adapted into a chunk-wise autoregressive model via teacher forcing, initialized for few-step generation via PDD~\citep{shaul2026parallel}. Finally, we leverage distribution matching distillation to obtain the target world model. Furthermore, we deploy inference optimizations covering computation graph fusion, low-bit quantization, KV caching, and a lightweight VAE to achieve real-time generation at 16 FPS on 8 H20 GPUs. See Appendix for more details.

\textbf{Evaluation Details.} To assess controllability, consistency, and visual fidelity, we benchmark all models on WBench~\citep{ying2026wbenchcomprehensivemultiturnbenchmark}. To systematically assess long-horizon geometric consistency, we introduce RevisitBench, an evaluation benchmark comprising 200 revisit trajectories following the protocol in \cite{sun2025worldplay}, which are curated from WBench and our self-collected validation sets. We quantify 2D visual consistency via LPIPS, PSNR, and SSIM, as well as 3D spatial consistency via MEt3R~\citep{asim2025met3r}. 
Additionally, we curate 145 samples from WBench and our interactive event dataset as test cases, which are held out from training, to evaluate model responsiveness to diverse interactive events. We deploy a vision-language model (VLM)~\citep{seed2_1} as an automated evaluator to score instruction adherence and execution accuracy.
We compare our approach against five interactive world models: WorldPlay~\citep{sun2025worldplay}, AlayaWorld~\citep{team2026alayaworld}, Lingbot-World-V2~\citep{gao2026infinite}, EchoWM~\citep{zhang2026echowm}, and Alaya-Evoke-Turbo~\citep{yin2026alaya}.

\subsection{Comparisons with Existing Methods}

\textbf{Quantitative Results.} Tab.~\ref{tab:long-term} compares our method with five representative interactive world models on the navigation split of WBench and RevisitBench. WorldPlay2 achieves the highest overall average score of 83.1, surpassing the previous state-of-the-art baseline, Alaya-Evoke-Turbo~\citep{yin2026alaya}, by 1.1 points. Notably, WorldPlay2 demonstrates a pronounced advantage in Interaction and Consistency. This empirically validates that our  factorized hybrid control interface, coupled with the co-design of memory and distillation, significantly enhances long-horizon consistency while maintaining precise navigation controllability. RevisitBench further evaluates long-horizon geometric consistency under loop-closure trajectories. Constrained by fixed context window lengths, Lingbot-World-V2~\citep{gao2026infinite} and EchoWM~\citep{zhang2026echowm} suffer from memory degradation, failing to preserve geometric consistency. Although WorldPlay~\citep{sun2025worldplay} incorporates camera poses for historical context retrieval, it is inherently susceptible to compounding camera pose drift as rollouts progress, making it challenging to retrieve accurate context. AlayaWorld~\citep{team2026alayaworld} and Alaya-Evoke-Turbo~\citep{yin2026alaya} rely on explicit 3D representations to maintain memory, but suffer from metric scale ambiguities across different chunks. Such scale discrepancies severely impede fine-grained control and inevitably induce geometric inconsistencies. In contrast, WorldPlay2 maintains robust long-horizon geometric consistency without relying on error-prone retrieval or sensitive explicit 3D representations. Furthermore, as evidenced by the results, Stable Forcing substantially improves performance, directly demonstrating its effectiveness.

% Furthermore, 从结果中可以看出Stable Forcing 极大的 enhances generation quality, directly demonstrating its effectiveness.

\begin{wraptable}{r}{0.52\textwidth}
\centering
\vspace{-20pt}
\caption{Quantitative comparison on responsiveness to interactive events. \textit{Abbreviations: EO: Environment and Object change, CI: Complex Interaction.}} \label{tab:interaction_compare}
% \vspace{-1pt}
\scriptsize
  \begin{tabular*}{\linewidth}{@{\extracolsep{\fill}}lccc@{}}
  \toprule
  \textbf{Method} & \textbf{Avg.} $\uparrow$ & \textbf{EO.} $\uparrow$ & \textbf{CI.} $\uparrow$ \\
  \midrule
  WorldPlay~\citep{sun2025worldplay}         & 43.3 & 47.9 & 38.7 \\
  AlayaWorld~\citep{team2026alayaworld}        & 39.4 & 48.3 & 30.5 \\
  Lingbot-World-V2~\citep{gao2026infinite}   & 52.5 & 61.5 & 43.4 \\
  EchoWM~\citep{zhang2026echowm}            & 43.8 & 56.3 & 31.2 \\
  Alaya-Evoke-Turbo~\citep{yin2026alaya}  & 40.7 & 45.7 & 35.7 \\
  Ours & \textbf{74.7} & \textbf{79.2} & \textbf{70.1} \\
  \bottomrule
  \end{tabular*}
  \vspace{-2mm}
\end{wraptable}
As shown in Tab.~\ref{tab:interaction_compare}, we evaluate responsiveness across diverse interactive events. Although Lingbot-World-V2\citep{gao2026infinite} and AlayaWorld~\citep{team2026alayaworld} leverage chunk-wise captions to accommodate diverse controls, they fail to disentangle underlying world states. This entangles content factors with interactive events, making it challenging for them to capture precise correspondences between textual descriptions and visual dynamics, particularly in complex interactions. In contrast, our model explicitly factorizes the world state via the factorized hybrid control interface, yielding superior interactive fidelity.

\textbf{Qualitative Results.} Fig.~\ref{fig:main_results} presents the qualitative comparisons with baselines. Due to the scarcity of interactive event datasets and reliance on entangled control conditioning, WorldPlay~\citep{sun2025worldplay} and AlayaWorld~\citep{team2026alayaworld} fail to respond accurately to diverse interactive commands. Meanwhile, Lingbot-World-V2~\citep{gao2026infinite} and EchoWM~\citep{zhang2026echowm} struggle to maintain long-horizon geometric consistency as a consequence of memory decay inherent in their sliding-window designs. Furthermore, their lack of factorized representations between background scenes and foreground characters impedes precise and smooth locomotion, such as airplane turning and entity centering. Although Alaya-Evoke-Turbo~\citep{yin2026alaya} can generate long sequences, its dependence on explicit 3D representations often introduces severe temporal flickering and visual artifacts, such as ghosting and duplicate entity appearances. In contrast, our approach reliably executes diverse interaction events, enables fine-grained and fluid control over different characters, and preserves long-horizon consistency, highlighting the efficacy and superiority of our method. Please refer to the supplementary videos for comprehensive visualizations.

\subsection{Ablation Studies}

\begin{figure*}[t]
  \centering
  \includegraphics[width=\textwidth]{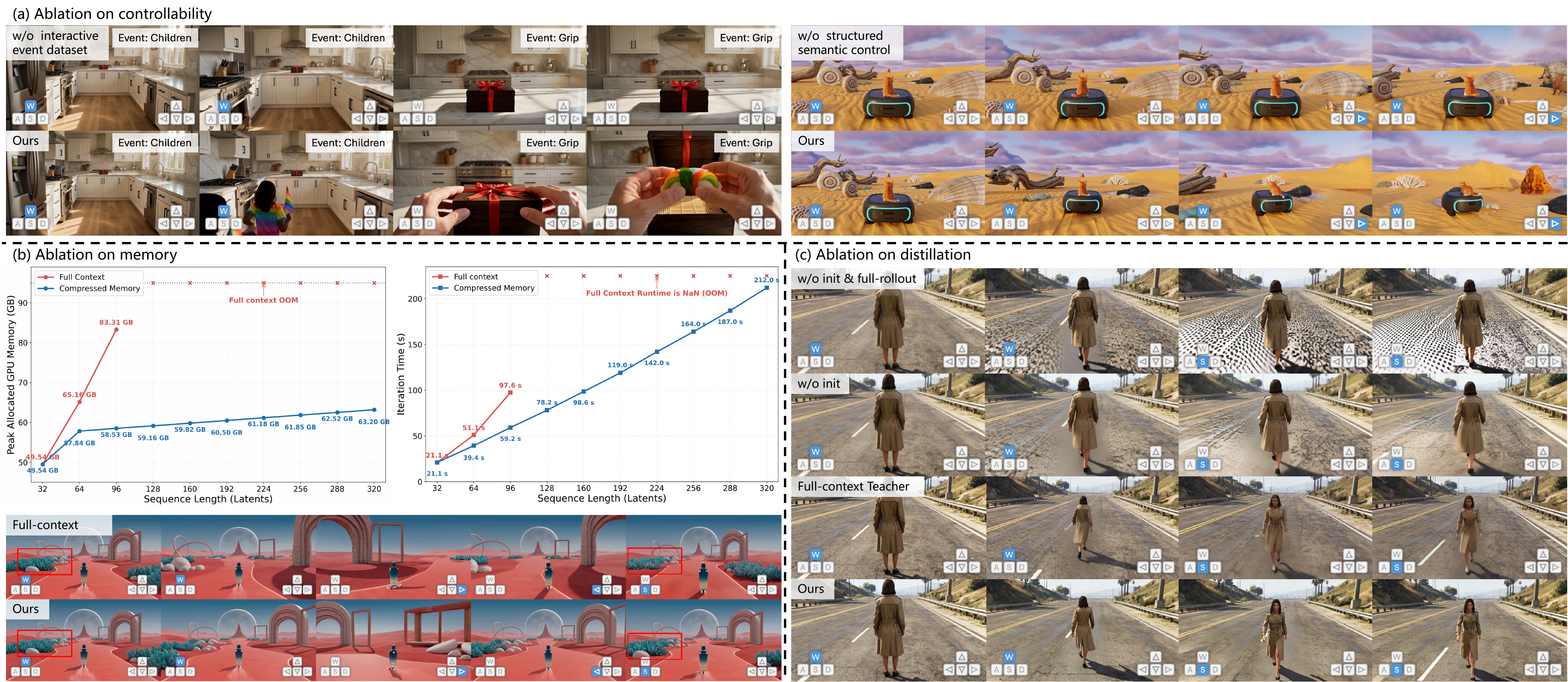}
  \vspace{-7mm}
  \caption{\textbf{(a) Ablation on controllability:} We verify the impact of interactive data scaling and structured semantic control. \textbf{(b) Ablation on memory:} We compare the GPU memory footprint and training time with the full-context baseline. For our method, we partition long sequences into multiple clips and compute sequentially within a single iteration. \textbf{(c) Ablation on distillation:} We analyze the components of Stable Forcing. Zoom in for details.
  }
  \label{fig:ablation}
  \vspace{-4mm}
\end{figure*}

\textbf{Controllability.} Fig.~\ref{fig:ablation}(a) presents the ablation study on controllability. When trained without the interactive event dataset, the model's interactive capabilities are constrained, failing to respond to semantic events. Incorporating even a small fraction of interactive data unlocks these capabilities, enabling coherent interactions spanning dynamic semantic events and navigation controls. Furthermore, omitting our structured semantic control causes the model to conflate intricate foreground characters with background scenes, compromising precise navigation controllability.

\textbf{Memory.} To validate the efficiency of our compressed memory, we compare its GPU memory footprint and training time with the full-context baseline across various video lengths, as summarized in Fig.~\ref{fig:ablation}(b). At the same sequence lengths, compressed memory substantially reduces both memory consumption and iteration time, enabling scalable training and distillation on longer sequences. As shown in Fig.~\ref{fig:ablation}(b), our compressed memory achieves comparable long-horizon geometric consistency to its full-context counterpart when trained on 96 latents, demonstrating its effectiveness.

\textbf{Distillation.} Fig.~\ref{fig:ablation}(c) demonstrates the stability and efficacy of Stable Forcing. We first ablate the full-rollout replay, omitting it leads to progressive quality degradation during long-horizon generation and eventually triggers mode collapse, as evidenced by the ground artifacts. Furthermore, we evaluate the impact of the PDD initialization. Employing PDD initialization effectively mitigates blurry outputs and grid-like artifacts, confirming the robustness of our method. Additionally, while distilling with the full-context teacher achieves competitive performance, it demands substantially longer training wall-clock time (261s vs. 167s per iteration). Furthermore, extending the full-context teacher to longer horizon distillation (\eg, 320 latents) triggers out-of-memory issues, whereas our method scales with superior efficiency.

\section{Conclusion and Limitations}
We present WorldPlay2, an interactive world model that achieves real-time responsiveness, versatile control, and long-horizon consistency. It significantly expands the interactive capabilities of world models, faithfully executing both navigation-oriented controls and semantic interactive events while maintaining geometric consistency over long horizons. Crucially, WorldPlay2 is designed with scalability at its core, which scales efficiently and stably as compute budgets and data volume expand. We hope WorldPlay2 serves as a crucial step toward advancements in embodied intelligence, spatial computing, and interactive entertainment.

\textbf{Limitations.} Despite the promising capabilities demonstrated by WorldPlay2, several challenges need further investigation. First, characters are still prone to gradual visual and semantic drift, occasionally failing to preserve strict identity consistency during long rollouts. Second, scaling our framework to infinite-horizon generation remains an open challenge. Maintaining both infinite rollout stability and long-term geometric consistency without error accumulation represents one of the most fundamental yet demanding frontiers in interactive world modeling.

\bibliography{iclr2027_conference}

@article{parker2025genie,
  title={Genie 3: A new frontier for world models},
  author={Parker-Holder, Jack and Fruchter, Shlomi and others},
  journal={Google DeepMind Blog},
  year={2025}
}

@inproceedings{sun2025worldplay,
  title={{WorldPlay}: Towards long-term geometric consistency for real-time interactive world modeling},
  author={Sun, Wenqiang and Zhang, Haiyu and Wang, Haoyuan and Wu, Junta and Wang, Zehan and Wang, Zhenwei and Wang, Yunhong and Zhang, Jun and Wang, Tengfei and Guo, Chunchao},
  booktitle={ICML},
  year={2026}
}

@article{wang2026worldcompass,
        title   = {WorldCompass: Reinforcement Learning for Long-Horizon World Models},
        author  = {Wang, Zehan and Wang, Tengfei and Zhang, Haiyu and Zuo, Xuhui and Wu, Junta and Wang, Haoyuan and Sun, Wenqiang and Wang, Zhenwei and Cao, Chenjie and Zhao, Hengshuang and Guo, Chunchao and Zhao, Zhou},
        booktitle={ICML},
        year    = {2026}
}

@article{he2025matrix,
  title={{Matrix-Game 2.0}: An open-source real-time and streaming interactive world model},
  author={He, Xianglong and Peng, Chunli and Liu, Zexiang and Wang, Boyang and Zhang, Yifan and Cui, Qi and Kang, Fei and Jiang, Biao and An, Mengyin and Ren, Yangyang and others},
  journal={arXiv preprint arXiv:2508.13009},
  year={2025}
}

@article{team2026advancing,
  title={Advancing open-source world models},
  author={Team, Robbyant and Gao, Zelin and Wang, Qiuyu and Zeng, Yanhong and Zhu, Jiapeng and Cheng, Ka Leong and Li, Yixuan and Wang, Hanlin and Xu, Yinghao and Ma, Shuailei and others},
  journal={arXiv preprint arXiv:2601.20540},
  year={2026}
}

@misc{happyoyster,
 author = {Alibaba},
 title = {{HappyOyster}},
 note = {\url{https://www.happyoyster.com}},
 year = 2026
}

@article{hong2025relic,
  title={{RELIC}: Interactive video world model with long-horizon memory},
  author={Hong, Yicong and Mei, Yiqun and Ge, Chongjian and Xu, Yiran and Zhou, Yang and Bi, Sai and Hold-Geoffroy, Yannick and Roberts, Mike and Fisher, Matthew and Shechtman, Eli and others},
  journal={arXiv preprint arXiv:2512.04040},
  year={2025}
}

@article{xu2026wonder,
  title={Wonder: Video World Model Done Better},
  author={Xu, Jiacong and Jiang, Hanwen and Shu, Zhixin and Sunkavalli, Kalyan and Patel, Vishal M and Mei, Yiqun},
  journal={arXiv preprint arXiv:2607.26037},
  year={2026}
}

@article{jiang2026abot,
  title={{ABot-World-0}: Infinite Interactive World Rollout on a Single Desktop GPU},
  author={Jiang, Fan and Sun, Zhaoxu and Wang, Mengchao and Zhu, Ziyu and Wang, Chiyu and Zhang, Yunpeng and Liu, Wenlin and Wang, Yun and Zheng, Xue and Sun, Rui and others},
  journal={arXiv preprint arXiv:2607.19191},
  year={2026}
}

@article{wan2025wan,
  title={Wan: Open and advanced large-scale video generative models},
  author={Wan, Team and Wang, Ang and Ai, Baole and Wen, Bin and Mao, Chaojie and Xie, Chen-Wei and Chen, Di and Yu, Feiwu and Zhao, Haiming and Yang, Jianxiao and others},
  journal={arXiv preprint arXiv:2503.20314},
  year={2025}
}

@article{team2026alayaworld,
  title={{AlayaWorld}: Interactive Long-Horizon World Modeling--Full Technical Report},
  author={Team, AlayaWorld and Zhang, Kaipeng and Li, Chuanhao and Zhan, Yifan and Ge, Yongtao and Yin, Yuanyang and Tan, Jiaming and He, Kang and Fan, Liaoyuan and Zhai, Mingliang and others},
  journal={arXiv preprint arXiv:2607.18367},
  year={2026}
}

@article{wu2025hunyuanvideo,
  title={Hunyuanvideo 1.5 technical report},
  author={Wu, Bing and Zou, Chang and Li, Changlin and Huang, Duojun and Yang, Fang and Tan, Hao and Peng, Jack and Wu, Jianbing and Xiong, Jiangfeng and Jiang, Jie and others},
  journal={arXiv preprint arXiv:2511.18870},
  year={2025}
}

@misc{veo,
 author = {Google Deepmind},
 title = {Veo3 video model},
 note = {\url{https://deepmind.google/models/veo/}},
 year = 2025
}

@misc{seed2_1,
 author = {ByteDance Seed},
 title = {Seed 2.1},
 note = {\url{https://seed.bytedance.com/en/seed2_1}},
 year = 2025
}

@article{agarwal2026cosmos,
  title={Cosmos 3: Omnimodal world models for physical ai},
  author={Agarwal, Niket and Ali, Arslan and Allen, Jon and Antolini, Martin and Aubame, Adeline and Azzolini, Alisson and Bai, Junjie and Bala, Maciej and Balaji, Yogesh and Bapst, Josh and others},
  journal={arXiv preprint arXiv:2606.02800},
  year={2026}
}

@article{brooks2024video,
  title={Video generation models as world simulators},
  author={Brooks, Tim and Peebles, Bill and Holmes, Connor and DePue, Will and Guo, Yufei and Jing, Leo and Schnurr, David and Taylor, Joe and Luhman, Troy and Luhman, Eric and others},
  journal={OpenAI Blog},
  volume={1},
  number={8},
  pages={1},
  year={2024}
}

@article{wiedemer2025video,
  title={Video models are zero-shot learners and reasoners},
  author={Wiedemer, Thadd{\"a}us and Li, Yuxuan and Vicol, Paul and Gu, Shixiang Shane and Matarese, Nick and Swersky, Kevin and Kim, Been and Jaini, Priyank and Geirhos, Robert},
  journal={arXiv preprint arXiv:2509.20328},
  year={2025}
}

@article{team2025evaluating,
  title={Evaluating gemini robotics policies in a veo world simulator},
  author={Team, Gemini Robotics and Choromanski, Krzysztof and Devin, Coline and Du, Yilun and Dwibedi, Debidatta and Gao, Ruiqi and Jindal, Abhishek and Kipf, Thomas and Kirmani, Sean and Leal, Isabel and others},
  journal={arXiv preprint arXiv:2512.10675},
  year={2025}
}

@article{gao2026infinite,
  title={Infinite Worlds with Versatile Interactions},
  author={Gao, Zelin and Wang, Qiuyu and Zhu, Jiapeng and Chen, Jingye and Liu, Zichen and Bai, Qingyan and Wang, Jiahao and Yuan, Yufeng and Wang, Hanlin and Lu, Yichong and others},
  journal={arXiv preprint arXiv:2607.07534},
  year={2026}
}

@article{zhu2026sana,
  title={{SANA-WM}: Efficient minute-scale world modeling with hybrid linear diffusion transformer},
  author={Zhu, Haoyi and Liu, Haozhe and Zhao, Yuyang and Ye, Tian and Chen, Junsong and Yu, Jincheng and He, Tong and Han, Song and Xie, Enze},
  journal={arXiv preprint arXiv:2605.15178},
  year={2026}
}

@article{shaul2026parallel,
  title={Parallel Decoding Distillation for Fast Image and Video Generation},
  author={Shaul, Neta and Liu, Chao and Vahdat, Arash and Berner, Julius},
  journal={arXiv preprint arXiv:2607.26004},
  year={2026}
}

@article{huang2026self,
  title={{Self Forcing}: Bridging the train-test gap in autoregressive video diffusion},
  author={Huang, Xun and Li, Zhengqi and He, Guande and Zhou, Mingyuan and Shechtman, Eli},
  journal={NeurIPS},
  volume={38},
  pages={167283--167308},
  year={2026}
}

@inproceedings{lipman2022flow,
  title={Flow matching for generative modeling},
  author={Lipman, Yaron and Chen, Ricky TQ and Ben-Hamu, Heli and Nickel, Maximilian and Le, Matt},
  booktitle={ICLR},
  year={2023}
}

@inproceedings{yu2025context,
  title={Context as memory: Scene-consistent interactive long video generation with memory retrieval},
  author={Yu, Jiwen and Bai, Jianhong and Qin, Yiran and Liu, Quande and Wang, Xintao and Wan, Pengfei and Zhang, Di and Liu, Xihui},
  booktitle={SIGGRAPH Asia},
  pages={1--11},
  year={2025}
}

@article{xiao2026worldmem,
  title={{WorldMem}: Long-term consistent world simulation with memory},
  author={Xiao, Zeqi and Lan, Yushi and Zhou, Yifan and Ouyang, Wenqi and Yang, Shuai and Zeng, Yanhong and Pan, Xingang},
  journal={NeurIPS},
  volume={38},
  pages={49632--49652},
  year={2026}
}

@article{zhang2025matrix,
  title={{Matrix-Game}: Interactive world foundation model},
  author={Zhang, Yifan and Peng, Chunli and Wang, Boyang and Wang, Puyi and Zhu, Qingcheng and Kang, Fei and Jiang, Biao and Gao, Zedong and Li, Eric and Liu, Yang and others},
  journal={arXiv preprint arXiv:2506.18701},
  year={2025}
}

@article{wang2026matrix,
  title={{Matrix-Game 3.0}: Real-time and streaming interactive world model with long-horizon memory},
  author={Wang, Zile and Liu, Zexiang and Li, Jiaxing and Huang, Kaichen and Xu, Baixin and Kang, Fei and An, Mengyin and Wang, Peiyu and Jiang, Biao and Wei, Yichen and others},
  journal={arXiv preprint arXiv:2604.08995},
  year={2026}
}

@article{li2025hunyuan,
  title={{Hunyuan-GameCraft}: High-dynamic interactive game video generation with hybrid history condition},
  author={Li, Jiaqi and Tang, Junshu and Xu, Zhiyong and Wu, Longhuang and Zhou, Yuan and Shao, Shuai and Yu, Tianbao and Cao, Zhiguo and Lu, Qinglin},
  journal={arXiv preprint arXiv:2506.17201},
  volume={2},
  number={3},
  pages={6},
  year={2025}
}

@article{tang2025hunyuan,
  title={{Hunyuan-GameCraft-2}: Instruction-following interactive game world model},
  author={Tang, Junshu and Liu, Jiacheng and Li, Jiaqi and Wu, Longhuang and Yang, Haoyu and Zhao, Penghao and Gong, Siruis and Yuan, Xiang and Shao, Shuai and Zhang, Linfeng and others},
  journal={arXiv preprint arXiv:2511.23429},
  year={2025}
}

@article{mao2025yume,
  title={Yume: An interactive world generation model},
  author={Mao, Xiaofeng and Lin, Shaoheng and Li, Zhen and Li, Chuanhao and Peng, Wenshuo and He, Tong and Pang, Jiangmiao and Chi, Mingmin and Qiao, Yu and Zhang, Kaipeng},
  journal={arXiv preprint arXiv:2507.17744},
  year={2025}
}

@inproceedings{mao2026yume1,
  title={Yume1.5: A text-controlled interactive world generation model},
  author={Mao, Xiaofeng and Li, Zhen and Li, Chuanhao and Xu, Xiaojie and Ying, Kaining and Zhang, Kaipeng},
  booktitle={CVPR},
  pages={7752--7761},
  year={2026}
}

@article{team2026dreamx,
  title={{DreamX-World 1.0}: A General-Purpose Interactive World Model},
  author={Team, DreamX and Bai, Yancheng and Chen, Rui and Chu, Xiangxiang and Dang, Rujing and Dou, Hao and Gao, Bingjie and Gu, Qiwen and Hong, Siyu and Lei, Jiachen and others},
  journal={arXiv preprint arXiv:2606.16993},
  year={2026}
}

@article{team2026inspatio,
  title={Inspatio-world: A real-time 4d world simulator via spatiotemporal autoregressive modeling},
  author={Team, InSpatio and Shen, Donghui and Zhang, Guofeng and Liu, Haomin and Ji, Haoyu and Bao, Hujun and Zhai, Hongjia and Liu, Jialin and Guo, Jing and Wang, Nan and others},
  journal={arXiv preprint arXiv:2604.07209},
  year={2026}
}

@article{geng2026mean,
  title={Mean flows for one-step generative modeling},
  author={Geng, Zhengyang and Deng, Mingyang and Bai, Xingjian and Kolter, Zico and He, Kaiming},
  journal={NeurIPS},
  volume={38},
  pages={75460--75482},
  year={2026}
}

@inproceedings{zheng2026large,
  title={Large scale diffusion distillation via score-regularized continuous-time consistency},
  author={Zheng, Kaiwen and Wang, Yuji and Ma, Qianli and Chen, Huayu and Zhang, Jintao and Balaji, Yogesh and Chen, Jianfei and Liu, Ming-Yu and Zhu, Jun and Zhang, Qinsheng},
  booktitle={ICLR},
  pages={2582--2603},
  year={2026}
}

@article{gu2026anyflow,
  title={{AnyFlow}: Any-step video diffusion model with on-policy flow map distillation},
  author={Gu, Yuchao and Fang, Guian and Jiang, Yuxin and Mao, Weijia and Han, Song and Cai, Han and Shou, Mike Zheng},
  journal={arXiv preprint arXiv:2605.13724},
  year={2026}
}

@inproceedings{wang2026transition,
  title={Transition models: Rethinking the generative learning objective},
  author={Wang, Zidong and Zhang, Yiyuan and Yue, Xiaoyu and Yue, Xiangyu and Li, Yangguang and Ouyang, Wanli and Bai, Lei},
  booktitle={CVPR},
  pages={29178--29189},
  year={2026}
}

@inproceedings{chen2026pi,
  title={{Pi-Flow}: Policy-based few-step generation via imitation distillation},
  author={Chen, Hansheng and Zhang, Kai and Tan, Hao and Guibas, Leonidas and Wetzstein, Gordon and Bi, Sai},
  booktitle={ICLR},
  pages={151521--151547},
  year={2026}
}

@inproceedings{yin2024one,
  title={One-step diffusion with distribution matching distillation},
  author={Yin, Tianwei and Gharbi, Micha{\"e}l and Zhang, Richard and Shechtman, Eli and Durand, Fredo and Freeman, William T and Park, Taesung},
  booktitle={CVPR},
  pages={6613--6623},
  year={2024},
}

@article{yin2024improved,
  title={Improved distribution matching distillation for fast image synthesis},
  author={Yin, Tianwei and Gharbi, Micha{\"e}l and Park, Taesung and Zhang, Richard and Shechtman, Eli and Durand, Fredo and Freeman, William T},
  journal={NeurIPS},
  volume={37},
  pages={47455--47487},
  year={2024}
}

@inproceedings{yin2025slow,
  title={From slow bidirectional to fast autoregressive video diffusion models},
  author={Yin, Tianwei and Zhang, Qiang and Zhang, Richard and Freeman, William T and Durand, Fredo and Shechtman, Eli and Huang, Xun},
  booktitle={CVPR},
  pages={22963--22974},
  year={2025},
}

@article{zheng2026causal,
  title={{Causal-rCM}: A Unified Teacher-Forcing and Self-Forcing Open Recipe for Autoregressive Diffusion Distillation in Streaming Video Generation and Interactive World Models},
  author={Zheng, Kaiwen and He, Guande and Zhao, Min and Zhang, Jintao and Chen, Huayu and Chen, Jianfei and Lin, Chen-Hsuan and Liu, Ming-Yu and Zhu, Jun and Ma, Qianli},
  journal={arXiv preprint arXiv:2606.25473},
  year={2026}
}

@inproceedings{zhu2026causal,
  title={{Causal Forcing}: Autoregressive diffusion distillation done right for high-quality real-time interactive video generation},
  author={Zhu, Hongzhou and Zhao, Min and He, Guande and Su, Hang and Li, Chongxuan and Zhu, Jun},
  booktitle={ICML},
  year={2026},
}

@inproceedings{zhang2025tinyhistory,
  title={{TinyHistory}: Lightweight Video History Embeddings via Two-Stage Context Learning},
  author={Zhang, Lvmin and Cai, Shengqu and Li, Muyang and Zeng, Chong and Lu, Beijia and Rao, Anyi and Han, Song and Wetzstein, Gordon and Agrawala, Maneesh},
  booktitle={ECCV},
  year={2026}
}

@inproceedings{wang2026spatialvid,
  title={{SpatialVID}: A large-scale video dataset with spatial annotations},
  author={Wang, Jiahao and Yuan, Yufeng and Zheng, Rujie and Lin, Youtian and Gao, Jian and Chen, Lin-Zhuo and Bao, Yajie and Zeng, Chang and Zhou, Yanxi and Long, Xiao-Xiao and others},
  booktitle={CVPR},
  pages={42592--42603},
  year={2026}
}

@article{li2026sekai,
  title={Sekai: A video dataset towards world exploration},
  author={Li, Zhen and Li, Chuanhao and Mao, Xiaofeng and Lin, Shaoheng and Li, Ming and Zhao, Shitian and Xu, Zhaopan and Li, Xinyue and Feng, Yukang and Sun, Jianwen and others},
  journal={NeurIPS},
  volume={38},
  year={2026}
}

@article{team2023gemini,
  title={Gemini: a family of highly capable multimodal models},
  author={Team, Gemini and Anil, Rohan and Borgeaud, Sebastian and Alayrac, Jean-Baptiste and Yu, Jiahui and Soricut, Radu and Schalkwyk, Johan and Dai, Andrew M and Hauth, Anja and Millican, Katie and others},
  journal={arXiv preprint arXiv:2312.11805},
  year={2023}
}

@article{bai2025qwen3,
  title={{Qwen3-VL} technical report},
  author={Bai, Shuai and Cai, Yuxuan and Chen, Ruizhe and Chen, Keqin and Chen, Xionghui and Cheng, Zesen and Deng, Lianghao and Ding, Wei and Gao, Chang and Ge, Chunjiang and others},
  journal={arXiv preprint arXiv:2511.21631},
  year={2025}
}

@article{huang2025vipe,
  title={{ViPE}: Video pose engine for 3d geometric perception},
  author={Huang, Jiahui and Zhou, Qunjie and Rabeti, Hesam and Korovko, Aleksandr and Ling, Huan and Ren, Xuanchi and Shen, Tianchang and Gao, Jun and Slepichev, Dmitry and Lin, Chen-Hsuan and others},
  journal={arXiv preprint arXiv:2508.10934},
  year={2025}
}

@misc{gpt6,
 author = {OpenAI},
 title = {GPT 6 Astra},
 note = {\url{https://openai.com/index/gpt-6-astra/}},
 year = 2026
}

@article{dong2024flex,
  title={{Flex Attention}: A programming model for generating optimized attention kernels},
  author={Dong, Juechu and Feng, Boyuan and Guessous, Driss and Liang, Yanbo and He, Horace},
  journal={arXiv preprint arXiv:2412.05496},
  volume={2},
  number={3},
  pages={4},
  year={2024}
}

@article{zhang2024sageattention2,
  title={{SageAttention2}: Efficient attention with thorough outlier smoothing and per-thread int4 quantization},
  author={Zhang, Jintao and Huang, Haofeng and Zhang, Pengle and Wei, Jia and Zhu, Jun and Chen, Jianfei},
  journal={arXiv preprint arXiv:2411.10958},
  year={2024}
}

@misc{BoerBohan2025TAEHV,
  author = {Boer Bohan, Ollin},
  title = {{TAEHV}: Tiny AutoEncoder for Hunyuan Video},
  year = {2025},
  howpublished = {\url{https://github.com/madebyollin/taehv}},
}

@article{zhang2026echowm,
  title={{EchoWM}: Open and Enterable Omnimodal World Models},
  author={Zhang, Songchun and Li, Yaowei and Zhuang, Junhao and Jin, Weiyang and Wang, Haoyu and Lu, Xin and Sun, Yilang and Zhang, Shiyi and Li, Haoran and Ma, Xiaoxiao and others},
  journal={arXiv preprint arXiv:2608.23189},
  year={2026}
}

@article{ying2026wbenchcomprehensivemultiturnbenchmark,
  title={{WBench}: A Comprehensive Multi-turn Benchmark for Interactive Video World Model Evaluation},
  author={Ying, Kaining and Hu, Hengrui and Ren, Siyu and Li, Jiamu and Chen, Fengjiao and Wang, Ziwen and Cao, Xuezhi and Cai, Xunliang and Ding, Henghui},
  journal={arXiv preprint arXiv:2605.25874},
  year={2026}
}

@inproceedings{asim2025met3r,
  title={{MEt3R}: Measuring multi-view consistency in generated images},
  author={Asim, Mohammad and Wewer, Christopher and Wimmer, Thomas and Schiele, Bernt and Lenssen, Jan Eric},
  booktitle={CVPR},
  pages={6034--6044},
  year={2025}
}

@article{yin2026alaya,
  title={{Alaya-EVOKE}: From Linear-Scaling Supervision to Endless World},
  author={Yin, Yuanyang and Wang, Gongxuan and Zhan, Yifan and Li, Chuanhao and Zhang, Kaipeng and Zhao, Feng},
  journal={arXiv preprint arXiv:2608.13546},
  year={2026}
}

@article{hunyuanworld_2025,
    title = {{HunyuanWorld 1.0}: Generating Immersive, Explorable, and Interactive 3D Worlds from Words or Pixels},  
    author  = {Tencent HY World Team},
    year    = {2025},
    journal={arXiv preprint arXiv:2507.21809},
}

@article{hyworld2_2026,
    title   = {{HY-World 2.0}: A Multi-Modal World Model for Reconstructing, Generating, and Simulating 3D Worlds},
    author  = {Tencent HY World Team},
    year    = {2026},
    journal={arXiv preprint arXiv:2604.14268},
}
\bibliographystyle{iclr2027_conference}

\newpage
\appendix
\section*{Appendix}
\section{Discussion}
% 和现在sliding window这种方式的区别，insight
A predominant paradigm among concurrent interactive world models~\citep{gao2026infinite, jiang2026abot} relies on sliding-window mechanisms to facilitate autoregressive rollout over extended temporal horizons. While computationally tractable, this formulation fundamentally enforces a local assumption: the model's predictive distribution is heavily conditioned on adjacent frames, inherently ignoring distant observations. Consequently, such architectures struggle with long-horizon geometric consistency. In contrast, our framework departs from local receptive fields by designing an efficient memory mechanism with a stable distillation. This enables our model to anchor spatiotemporal invariants across long rollouts, maintaining high-fidelity geometric and semantic consistency. Despite these gains, we candidly acknowledge the boundaries of our approach. Scaling our framework to unbounded, infinite-horizon generation remains an open challenge. We posit that achieving both infinite-horizon rollout and long-term geometric persistence represents one of the most fundamental frontiers in interactive world modeling. Furthermore, recent coding agents~\citep{gpt6} have demonstrated a remarkable ability to synthesize spatially coherent 3D environments via programmatic generation. This milestone signals a profound paradigm shift, transitioning agents from static textual domains to dynamic embodied multi-modal domains. We believe that integrating such intelligence into generative world models represents an exceptionally promising frontier.

\section{Dataset}
\subsection{Spatial Navigation Dataset}
% gamerecord，UE，abot, SpatialVid, Sekai
As detailed in Tab.~\ref{tab:dataset_composition}, our spatial navigation dataset comprises three complementary categories. First, to capture real-world physical dynamics and complex photorealistic textures, we curate high-quality subsets from SpatialVID~\citep{wang2026spatialvid} and Sekai~\citep{li2026sekai}. Specifically, we curate videos captured in open unobstructed environments with smooth navigation trajectories. Second, to reinforce long-term geometric consistency under complex camera trajectories, we design a dedicated loop-closure rendering pipeline in UE. The synthesized camera trajectories are strictly self-symmetric, \ie, the second half of the sequence precisely reverses the camera path of the first half. During the first half exploration phase, multiple navigation actions are executed with randomized durations ranging from 1s to 5s. Finally, to broaden trajectory diversity and improve generalization, we incorporate open-source gameplay datasets~\citep{jiang2026abot} and scale the simulation recording dataset. This substantially enriches the coverage of game genres, dynamic environments, and third-person characters.

\subsection{Interactive Event Dataset}
\begin{figure*}[t]
  \centering
  \includegraphics[width=\textwidth]{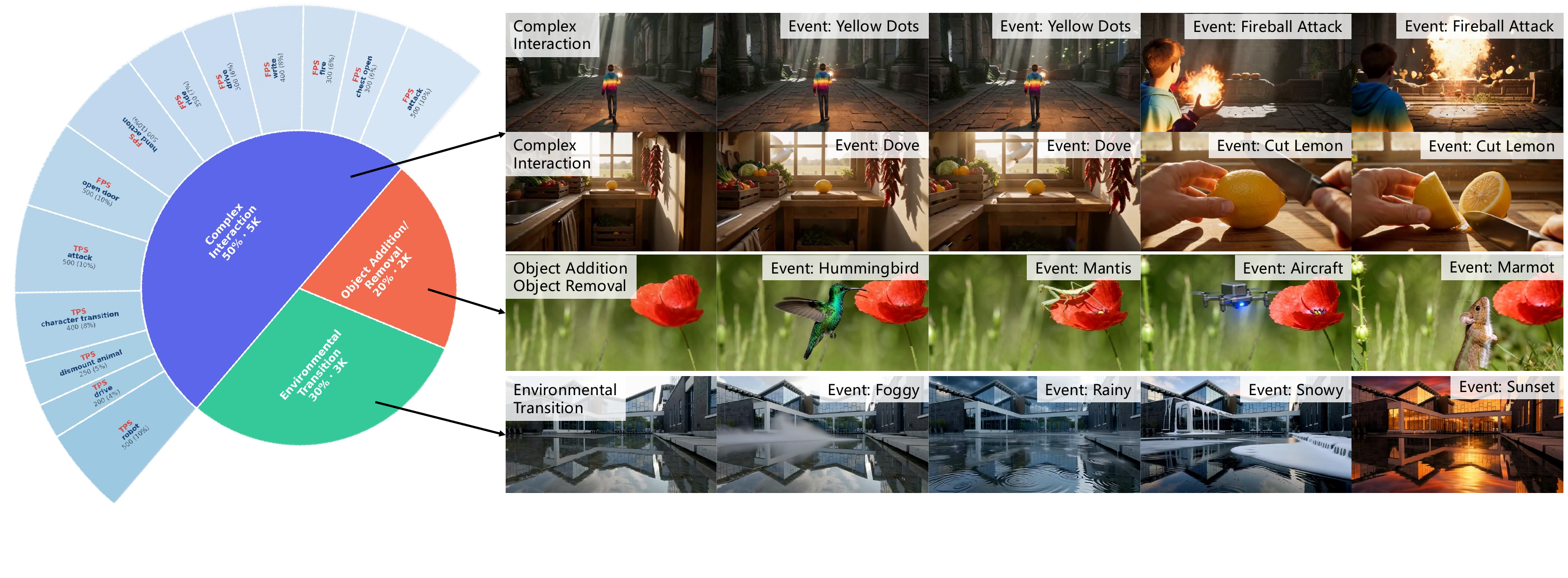}
  \vspace{-12mm}
  \caption{\textbf{Overview of our interactive event dataset.} The dataset encompasses three categories, including complex interaction, environmental transition, and object addition/removal. \textit{Left}: The composition of the dataset. \textit{Right}: Representative visual examples for each category.}
  \label{fig:event_dataset}
\end{figure*}

% 为什么我们的动作比较简单，3类，
Our interactive event dataset comprises three categories: complex interaction, environmental transition, and object addition/removal, as detailed in Tab.~\ref{tab:dataset_composition} and Fig.~\ref{fig:event_dataset}. For complex interaction, we aim to jointly capture intricate behavioral interactions alongside spatial navigation, spanning 13 distinct action categories across both first-person and third-person perspectives. For environmental transition, we collect diverse videos capturing dynamic weather shifts (\eg, clear skies transitioning to heavy rain, fog, or snow), seasonal evolutions (\eg, summer lushness shifting to autumnal foliage or winter snowscapes), and holistic stylistic variations (\eg, day-to-night lighting and artistic rendering styles). For object addition and removal, our dataset encompasses the emergence and disappearance of dynamic agents such as animals, flying crafts (\eg, drones and aircraft), vehicles, tools, boxes, and handheld items.

\subsection{Data Processing}
% 包括vipe，caption怎么打标，怎么过滤
Our data processing pipeline consists of three stages: structured semantic captioning, spatial navigation action extraction, and quality filtering.

\textbf{Structured Semantic Captioning.} In alignment with our structured semantic control, each video clip requires a disentangled textual description. To this end, we utilize VLMs~\citep{team2023gemini, bai2025qwen3} to parse the visual information into structured captions using the following template.
\begin{AcademicBox}[Structured Semantic Captioning Template]
\textbf{Role:} You are a professional video annotator specializing in interactive world models. \\[2pt]
\textbf{Core Task:} You will be given a video. Your task is to describe it and output a single JSON object. Follow these rules exactly. \\[2pt]
\textbf{Important:} This video has been pre-labeled as: {PERSPECTIVE}. Trust this label and follow the corresponding perspective handling rule below. \\ [4pt]
\textbf{--- Rules ---} \\ [2pt]
\textbf{Language:} Write all field values in English. \\[2pt]
\textbf{Output:} Return ONLY one valid JSON object with exactly four keys: \textit{scene}, \textit{character}, \textit{UI}, \textit{dynamic\_interactions}. No extra text, no markdown code fences, no commentary. Do not describe camera motion, camera rotation, camera shake, zooming, viewpoint movement, gameplay controls. Be specific, concrete, and visually grounded. Do not invent details that cannot be seen. \\[2pt]
\textbf{Perspective handling (affects the \textit{character} field only):} TPS: Describe the visual appearance of any visible player-controlled subject (humanoid, vehicle, or creature) under \textit{character}. FPS: Set \textit{character} to "None". \\[2pt]
\textbf{Scene:} A unified descriptor encompassing visible terrain/objects, the environmental setting (e.g., forest, urban, room), and aesthetic styles (e.g., lighting, rendering scheme). Dynamic actions and transient events are prohibited. \\[2pt]
\textbf{Character:} An appearance-only descriptor of the controlled agent, strictly omitting transient poses, gestures, or kinetic activities. It details four morphological dimensions: agent taxonomy \& build, surface styling, equipped accessories, and active mounts. \\[2pt]
\textbf{UI:} A unified spatial layout descriptor of all visible on-screen interface elements—including HUD components, status meters, minimaps, crosshairs, and text overlays—along with their designated screen positions, defaulting to "None" when no graphical interface is present. \\[2pt]
\textbf{dynamic\_interactions}: A chronological array of 5s intervals (start\_time, end\_time, event) cataloging visually grounded causal dynamics. It prioritizes salient physical engagements, hand gestures, contact transitions (approach, manipulation, release), and resulting state changes. Camera motions, static descriptions, and speculative actions are strictly excluded; inactive segments default to "None". \\[4pt]
\textbf{--- Output Format ---}
\begin{lstlisting}[language=json]
{
    "scene": "...",
    "character": "...",
    "UI": "...",
    "dynamic_interactions": [
      {
        "start_time": 0,
        "end_time": 5,
        "event": "..."
      },
      {
        "start_time": 5,
        "end_time": 10,
        "event": "..."
      }
    ]
}
\end{lstlisting}
\end{AcademicBox}

\textbf{Spatial Navigation Action Extraction.} We observe that directly applying ViPE~\citep{huang2025vipe} to long-horizon video clips frequently suffers from severe trajectory drift and scale collapse. To mitigate this degradation, we partition each long video into overlapping clips of 150 frames with a 10-frame temporal overlap. Camera poses are estimated independently for each sub-clip and subsequently stitched together by aligning metric scales using the overlapping windows. From the reconstructed camera poses, we derive frame-aligned action controls by decomposing the motion into discrete longitudinal and lateral translations alongside continuous pitch and yaw angles. Our internal gameplay recordings are captured under constant angular velocities per game. Leveraging this property, we compute the directional angular velocities offline for each game and adopt their empirical medians as the calibrated rotation rates along the corresponding axes.

\textbf{Quality Filtering.} Our quality filtering pipeline operates along two distinct dimensions: visual fidelity and action precision. Following~\cite{wu2025hunyuanvideo}, we employ a comprehensive visual quality assessment model coupled with an aesthetic scoring operator to evaluate video clips across five perceptual dimensions, \ie, sharpness, fine-detail retention, noise and compression artifacts, dynamic range, and aesthetic appeal, systematically filtering out low-quality candidates. Then, we evaluate the temporal smoothness and physical plausibility of the estimated camera poses for each video clip, filtering out abrupt camera jitter to ensure accurate action labels.

\section{More Implementation Details}

\begin{table}[t]
  \centering
  \caption{\textbf{Data organization.} We detail the data type, category, data source, clip counts, and their corresponding proportions in the training corpus.}
  \label{tab:dataset_composition}
  \vspace{2mm}

  {\footnotesize
  \renewcommand{\arraystretch}{1.15} % 调整行高

  \begin{tabularx}{\linewidth}{
     >{\raggedright\arraybackslash}p{0.10\linewidth} % Data Type
     |
      >{\raggedright\arraybackslash}p{0.28\linewidth} % Purpose
      >{\raggedright\arraybackslash}X                % Data Source
      >{\raggedleft\arraybackslash}p{0.07\linewidth} % Quantity
      >{\raggedleft\arraybackslash}p{0.06\linewidth} % Ratio
  }
      \toprule
      \textbf{Data Type}
      & \textbf{Category}
      & \textbf{Data Source}
      & \textbf{Quantity}
      & \textbf{Ratio} \\
      \midrule

      \multirow{5}{=}{Spatial \\ Navigation}
      & \multirow{2}{=}{Real-World Dynamics}
      & SpatialVID~\citep{wang2026spatialvid}
      & 30K & 4.22\% \\
      
       &
      & Sekai~\citep{li2026sekai}
      & 20K & 2.81\% \\

      & Synthetic 4D Scenes
      & UE Rendering & 50K & 7.04\% \\

      & \multirow{2}{=}{Simulation Dynamics}
      & ABot-World~\citep{jiang2026abot} & 30K & 4.22\% \\

      &
      & Gameplay Recording
      & 570K & 80.28\% \\

      \midrule

      \multirow{3}{=}{Interactive \\ Event}
      & Complex Interaction
      & \multirow{3}{*}{Internal Dataset}
      & 5K & 0.70\% \\
    
      & Environmental Transition
      &
      & 3K & 0.42\% \\
    
      & Object Addition/Removal
      &
      & 2K & 0.28\% \\

      \bottomrule
  \end{tabularx}
  }
\end{table}

\subsection{Architecture Details}

\begin{wrapfigure}{r}{0.38\textwidth}
  \centering
  \vspace{-60pt}
  \includegraphics[width=\linewidth]{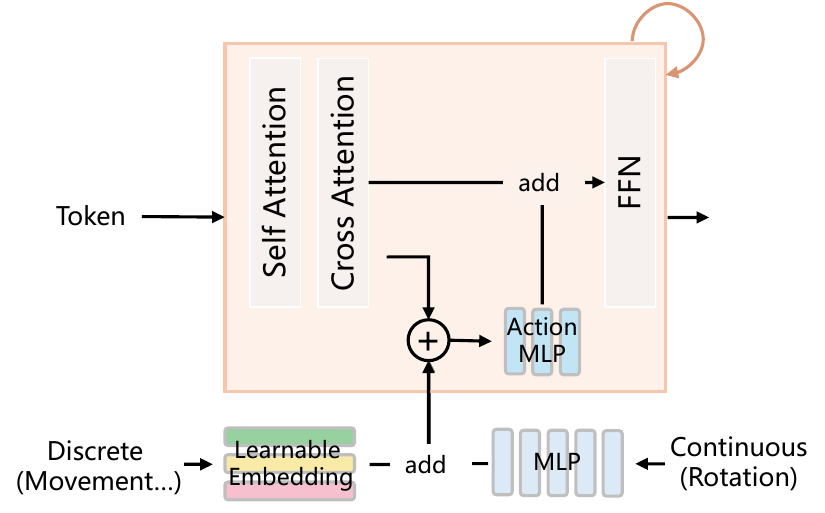}
  \vspace{-18pt}
  \caption{Illustration of our action module.}
  \label{fig:supp_arch}
  \vspace{-10pt}
\end{wrapfigure}
\textbf{Frame-aligned Action Module.} Since our frame-aligned actions contain both continuous camera rotations and discrete controls, we devise tailored encoding pathways as shown in Fig.~\ref{fig:supp_arch}. For continuous camera rotations, we employ an MLP to model varying angular velocities. For discrete controls, we adopt learnable embeddings to effectively capture concrete patterns. The two action embeddings are then added and injected before the FFN in each Transformer block. Interestingly, we observe that omitting explicit camera poses incurs no visible degradation in navigation control. Moreover, enforcing rigid camera trajectories makes it difficult to simulate character-centric rotations. In contrast, our design unleashes smoother, more fluid third-person character locomotion and delivers stronger generalization across diverse characters.

\begin{figure*}[t]
  \centering
  \includegraphics[width=\textwidth]{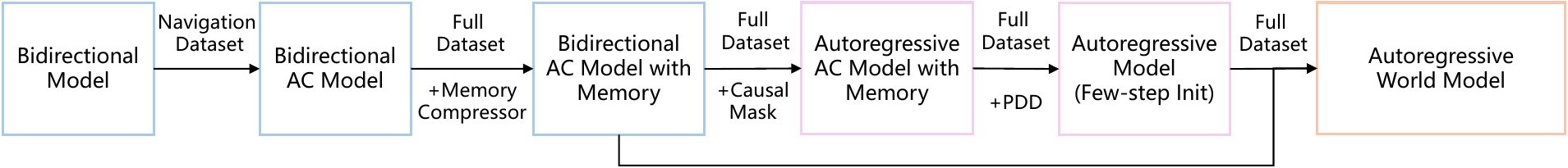}
  \vspace{-6mm}
  \caption{\textbf{Overview of our multi-stage training pipeline.} The bidirectional base model is progressively adapted into a streaming autoregressive model via action conditioning (AC), memory integration, and distillation.}
  \label{fig:training_pipeline}
  \vspace{-2mm}
\end{figure*}

\begin{algorithm}[t]
\caption{Stable Forcing (Full-rollout Replay and Efficient Score Evaluation)}
\label{alg:stable_forcing}
\begin{algorithmic}[1]
\REQUIRE Initialized student $N_\theta$;
real and fake score models $v_{\text{real}},v_{\text{fake}}$; the number of rollout chunks $M$;
denoising schedule $\{\sigma_d\}_{d=0}^{D}$;
\ENSURE Few-step autoregressive student.

\STATE $\mathcal{H} \gets \emptyset$; $\mathcal{X} \gets \emptyset$

\STATE \textbf{Stage 1: Full-rollout without gradients}
\FOR{$i = 1,\ldots,M$}
    \STATE $m_{<i} \gets \mathrm{CompressedMemory}(\mathcal{H})$
    \STATE $z_{i} \sim \mathcal{N}(0,I)$;
    $d_i \sim \mathrm{Uniform}(\{0,\ldots,D-1\})$
    \FOR{$d = 0,\ldots,D-1$}
        \IF{$d = d_i$}
            \STATE $\mathcal{R}_i \gets
            \mathrm{Snapshot}(z_{i},\sigma_{d},m_{<i},A_{\le i})$
        \ENDIF
        \STATE $z_{i} \gets
        \mathrm{Denoise}(z_{i}, N_{\theta}(z_{i},\sigma_{d},m_{<i},A_{\le i}))$
    \ENDFOR
    \STATE Append $\text{sg}(z_{i})$ to $\mathcal{X}$ and $\mathcal{H}$
\ENDFOR

\STATE \textbf{Stage 2: Efficient score evaluation without gradients}
\STATE Partition $\mathcal{X}$ into $B$ consecutive clips $\{x_b\}_{b=1}^{B}$
\FOR{$b = 1,\ldots,B$}
    \STATE Sample score timestep $\sigma$ and noise $\epsilon$
    \STATE $x_b^\sigma \gets \mathrm{AddNoise}(x_b,\sigma,\epsilon)$
    \STATE $g_b \gets \big[
    v_{\text{fake}}(x_b^\sigma,\sigma,m_{<b},A_{\le b})
    -v_{\text{real}}(x_b^\sigma,\sigma,m_{<b},A_{\le b})\big]$
\ENDFOR

\STATE \textbf{Stage 3: Gradient replay with gradients}
\FOR{$i = 1,\ldots,M$}
    \STATE $\widetilde{z}_i \gets
    \mathrm{CleanPrediction}(z_i,\sigma_{d_i},N_{\theta}(\mathcal{R}_{i}))$
    \STATE $\mathcal{L}_i \gets \text{MSE}(\widetilde{z}_i, \text{sg}(\widetilde{z}_i - g_i))$
    \STATE Backpropagate $\mathcal{L}_i$
\ENDFOR
\STATE Update $\theta$ once using the accumulated gradients
\STATE Update $v_{\text{fake}}$ following DMD2~\citep{yin2024improved}
\end{algorithmic}
\end{algorithm}

\textbf{History Compressor.} Our history compressor builds upon the design paradigm of TinyHistory~\citep{zhang2025tinyhistory}, combining 3D convolutions to compress spatiotemporal history with attention modules to enhance expressiveness. We further replace standard 3D convolutions with causal 3D convolutions. This design strictly prevents future information leakage, thereby adhering to the temporal causality of autoregressive generation.

\subsection{Training Details}
% 训练细节，table，超参数
Our multi-stage training pipeline is summarized in Fig.~\ref{fig:training_pipeline}. For the bidirectional model, each training sequence comprises 32 target latents with the associated memory context. For the AR model, the training sequence is formulated over 16 target latents, \ie, 4 chunks, along with their corresponding memory context. To optimize AR training throughput, we pad the variable-length memory contexts to a uniform sequence length, which enables efficient kernel execution via FlexAttention~\citep{dong2024flex}. Furthermore, both bidirectional and AR models are trained using a progressive curriculum, increasing the maximum data length to facilitate smooth convergence. During Stable Forcing, we leverage the AR model as the student, distilling it into 4 steps under the supervision of the bidirectional teacher model. Specifically, the student model performs self-rollout over a horizon of 320 latents, which the teacher partitions into 10 clips to efficiently compute scores. To compute backward passes over these long-horizon sequences, we adopt a gradient replay strategy analogous to RELIC~\citep{hong2025relic}, effectively reducing peak GPU memory footprints during backpropagation. Alg.~\ref{alg:stable_forcing} outlines the pseudocode of Stable Forcing.

\begin{table}[t]
  \centering
  \caption{\textbf{Inference speed improvements from system optimizations.} Latency is benchmarked as the average time per chunk across multiple generated chunks. Latency reductions are reported relative to the baseline.}
  \label{tab:inference_acceleration}

  \small
  \renewcommand{\arraystretch}{1.12}
  \setlength{\tabcolsep}{5pt}

  \begin{tabularx}{\linewidth}{
      >{\raggedright\arraybackslash}X
      >{\centering\arraybackslash}p{0.16\linewidth}
      >{\centering\arraybackslash}p{0.16\linewidth}
      >{\centering\arraybackslash}p{0.18\linewidth}
  }
      \toprule
      \textbf{Optimization}
      & \textbf{Latency (s)}
      & \textbf{Speedup}
      & \textbf{Reduction} \\
      \midrule

      Baseline
      & 4.138
      & $1.00{\times}$
      & -- \\

      $+$ QKRoPE
      & 4.103
      & $1.01{\times}$
      & 0.85\% \\

      $+$ QKVFusion
      & 4.077
      & $1.01{\times}$
      & 1.47\% \\

      $+$ FFNCompile
      & 4.074
      & $1.02{\times}$
      & 1.55\% \\

      $+$ Text Cache
      & 3.955
      & $1.05{\times}$
      & 4.42\% \\

      $+$ FP8 Quantization
      & 3.805
      & $1.09{\times}$
      & 8.05\% \\

      $-$ FSDP
      & 3.562
      & $1.16{\times}$
      & 13.92\% \\

      $+$ SageAttention~\citep{zhang2024sageattention2}
      & 3.432
      & $1.21{\times}$
      & 17.06\% \\

      $+$ LightVAE (Final)
      & \textbf{0.998}
      & \textbf{$4.15{\times}$}
      & 75.88\% \\

      \bottomrule
  \end{tabularx}
\end{table}

\subsection{Inference Details}
% 推理怎么加速
Complementing our algorithmic acceleration, we implement end-to-end systems optimizations across the entire inference pipeline, ultimately achieving a real-time streaming throughput of 16 FPS on 8 NVIDIA H20 GPUs as shown in Tab.~\ref{tab:inference_acceleration}. These optimizations include three complementary dimensions, \ie, computational graph fusion, quantization coupled with caching reuse, and lightweight VAE.

\textbf{Computational Graph Fusion.} We fuse the rotary position embeddings with the query-key linear projections (QKRoPE), combine query-key-value linear projections into a single linear projection (QKVFusion), and apply block-level compilation to the feed-forward networks (FFNCompile). 

\textbf{Quantization Coupled with Caching Reuse.} We incorporate FP8 quantization (FP8 Quantization) alongside SageAttention~\citep{zhang2024sageattention2} to further compress execution latency and peak memory footprints. Moreover, we cache the text embeddings (Text Cache), which bypasses redundant text encoding passes. Finally, we disable Fully Sharded Data Parallel (FSDP) during inference, thereby eliminating cross-device collective communication overheads.

\textbf{Lightweight VAE.} To overcome the severe latency bottleneck inherent in video VAE, we redesign and retrain a lightweight VAE grounded on TAE~\citep{BoerBohan2025TAEHV}.

\subsection{Evaluation Details}

To systematically evaluate long-horizon geometric consistency, we construct RevisitBench, comprising 50 10s and 150 30s test cases with loop-closure trajectories. Specifically, for each test case, we first randomly sample discrete movement and continuous rotation to generate the trajectory for the first half of the sequence. For the remaining duration, we invert the preceding action sequence, making the model revisit its original path. These rigorous loop-closure cases reliably evaluate the geometric consistency of world models.

As described in the main paper, to evaluate the responsiveness to diverse interactive events, we leverage the VLM as an automated evaluator to quantitatively verify instruction adherence and execution accuracy. Specifically, regarding environment and
object change, we assess whether the expected state transitions faithfully occur and reach completion, as well as whether these visual transformations semantically correspond to the given interactive controls. For complex interactions, we further examine whether the fine-grained interactive details strictly align with the user-specified control inputs. We employ the following prompt template.
\begin{AcademicBox}[Interactive Evaluation Template]
\textbf{Role:} You are an impartial evaluator of interactive world models. Evaluate how faithfully the video responds to the provided interactive control instructions, using only visual observations. \\[2pt]
\textbf{Inputs:} Video: \verb|<VIDEO>|; Interaction category: \verb|<EO or CI>|; Control instructions: \verb|<INSTRUCTIONS>|; Instruction timing: \verb|<FRAME RANGES>|; \\[2pt]
\textbf{Evaluate the following dimensions:} \\ [2pt]
\noindent\textbf{A. Instruction Adherence.} Assess whether the observed response matches the requested action, target, and attributes, without incorrect targets, reversed actions, or unintended substitutions. Assess whether the action or state transition reaches its intended endpoint. Distinguish completion from attempts or partial execution. \\ [1pt]
\noindent\textbf{B. Execution Accuracy.} Assess whether the requested action is visibly executed correctly. For \textit{EO}, verify that the intended state transition occurs with the specified direction and attributes. For \textit{CI}, verify the interaction and relevant details, including contact location, motion direction, manipulation, and the resulting relationship. Do not impose requirements beyond the instruction.

\textbf{--- Scoring ---}

Assign an integer from 0 to 4 to each dimension:

- 0: No fulfillment, or clear contradiction of the requirement.

- 1: Minimal fulfillment, with major errors or only a weak attempt.

- 2: Partial fulfillment, with substantial omissions or errors.

- 3: Mostly fulfilled, with minor errors or limited incompleteness.

- 4: Fully fulfilled, supported by clear visual evidence.

\textbf{--- Output Format ---}
\begin{lstlisting}[language=json]
{
    "category": "EO or CI",
    "evaluations": [
      {
        "instruction_id": 1,
        "instruction": "...",
        "scores": {
          "instruction_adherence": "3",
          "execution_accuracy": "4",
        },
        "reasoning": "Provide the evidence."
      }
    ]
}
\end{lstlisting}
\end{AcademicBox}

\begin{figure*}[t]
  \centering
  \includegraphics[width=\textwidth]{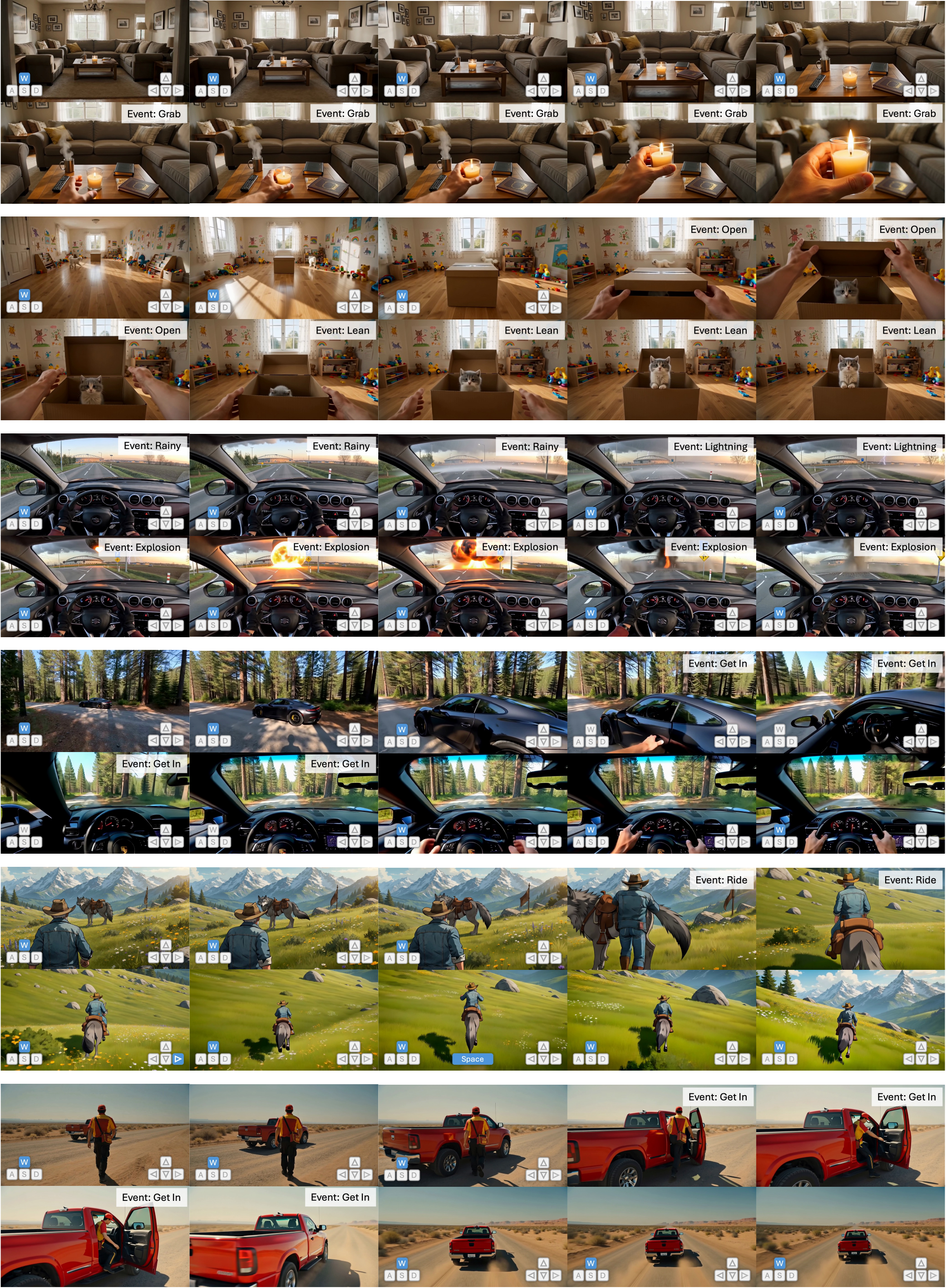}
  % \vspace{-2mm}
  \caption{\textbf{Qualitative visualizations on versatile controls.} WorldPlay2 supports a broad range of open-ended interactions, such as fine-grained hand-object manipulations (dexterous grasping and box opening), dynamic environmental transitions and anomaly events (weather shifts and explosion), and complex full-body interactions (getting in vehicles and riding animals).}
  \label{fig:supp_vis1}
\end{figure*}

\begin{figure*}[t]
  \centering
  \includegraphics[width=\textwidth]{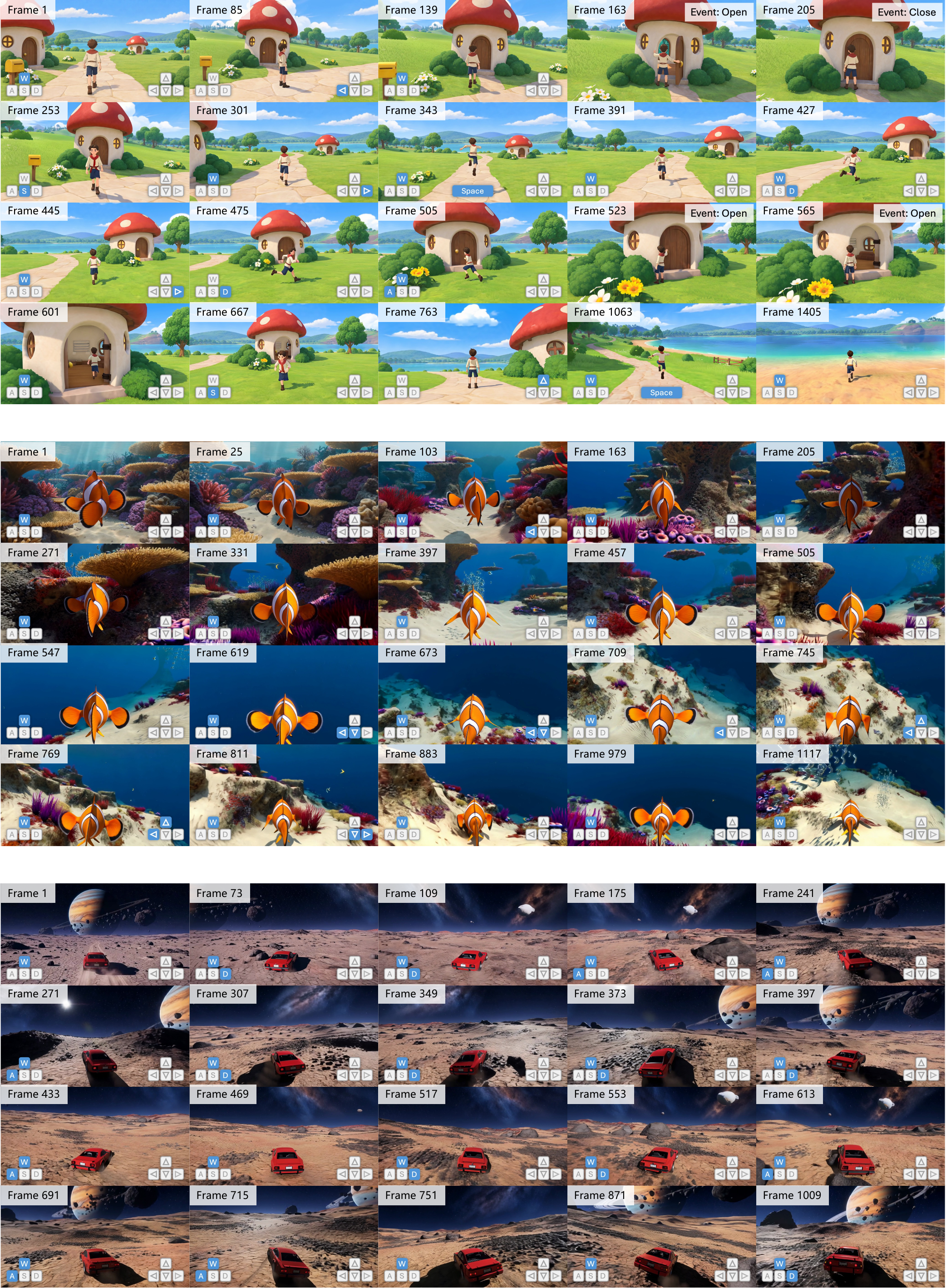}
  % \vspace{-2mm}
   \caption{\textbf{Qualitative visualizations on long-horizon consistency.} WorldPlay2 generalizes robustly across different characters. Crucially, it can execute intricate interactions seamlessly alongside navigation controls (first case), while faithfully preserving spatiotemporal consistency over long horizons.}
  \label{fig:supp_vis2}
\end{figure*}

\begin{figure*}[t]
  \centering
  \includegraphics[width=\textwidth]{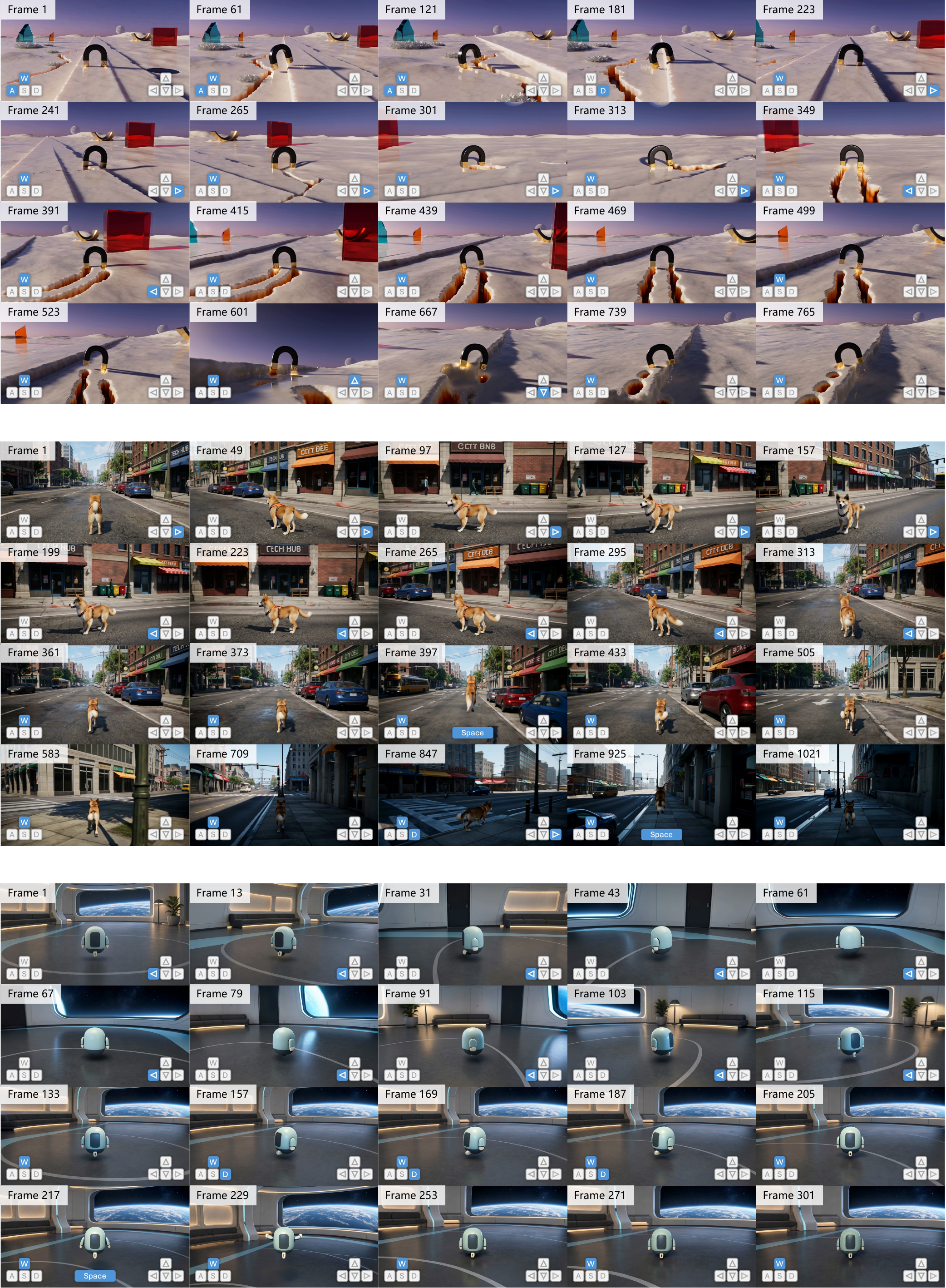}
  % \vspace{-2mm}
  \caption{\textbf{Qualitative visualizations on long-horizon consistency.} \textbf{Top:} WorldPlay2 maintains geometric consistency under compound navigation controls. \textbf{Middle:} It generates structurally plausible scene layouts and preserves spatial coherence over long horizons. \textbf{Bottom:} It preserves geometric consistency under $360^\circ$ camera rotations.}
  \label{fig:supp_vis3}
\end{figure*}

\section{More Results}

\subsection{More Visualizations} 

Fig.~\ref{fig:supp_vis1}, Fig.~\ref{fig:supp_vis2}, and Fig.~\ref{fig:supp_vis3} illustrate the comprehensive qualitative performance of WorldPlay2 across diverse environments, different characters, and versatile controls. WorldPlay2 not only accommodates intricate semantic interactions (\eg, dexterous grasping and box opening) alongside macro-level environmental transitions (\eg, weather shift and explosion), but also exhibits exceptional third-person character controllability. Specifically, it faithfully synthesizes smooth character motions that adhere strictly to the physical and kinematic constraints of diverse entities. Furthermore, it sustains robust geometric consistency and structural fidelity even under compound, multi-stage interactive controls.

\subsection{Quantitative Ablations}

\textbf{Controllability.} To validate the factorized hybrid control interface, we perform ablations on our bidirectional teacher model. As presented in Tab.~\ref{tab:ablation_event}, we first probe the role of the interactive event dataset on semantic responsiveness. We observe that even in the absence of the interactive event data, the model already exhibits moderate responsiveness. Consequently, introducing a small fraction of interactive data is highly sample-efficient, yielding substantial gains in semantic interactivity. Moreover, we assess navigation performance regarding structured semantic control as in Tab.~\ref{tab:ablation_semantic}. Ablating this module causes navigation conditioning to couple with other world states, inducing semantic entanglement that degrades navigation accuracy.
\begin{table}[t]
  \centering
  \setlength{\parskip}{0pt}

  % ========================================================
  % 第一行：Caption 顶部对齐
  % ========================================================
  \noindent
  \begin{minipage}[t]{0.49\linewidth}
  \vspace{0pt}
  \caption{Quantitative comparison on interactive event dataset.}
  \label{tab:ablation_event}
  \end{minipage}\hfill%
  \begin{minipage}[t]{0.49\linewidth}
  \vspace{0pt}
  \caption{Quantitative comparison on structured semantic control.}
  \label{tab:ablation_semantic}
  \end{minipage}

  % 去掉两排盒子之间自动添加的行间距
  \par\nointerlineskip
  \vspace{2pt}

  % ========================================================
  % 第一行：表格底部对齐
  % ========================================================
  \noindent
  \begin{minipage}[b]{0.49\linewidth}
  \scriptsize
  \setlength{\tabcolsep}{2pt}
  \renewcommand{\arraystretch}{1.1}

  \begin{tabular*}{\linewidth}[b]
      {@{\extracolsep{\fill}}lccc@{}}
  \toprule
  \textbf{Method}
  & \textbf{Avg.} $\uparrow$
  & \textbf{EO.} $\uparrow$
  & \textbf{CI.} $\uparrow$ \\
  \midrule
  w/o interactive event dataset
  & 46.7 & 57.5 & 35.8 \\
  Ours (Bidirectional)
  & \textbf{84.0}
  & \textbf{86.8}
  & \textbf{81.2} \\
  \bottomrule
  \end{tabular*}
  \end{minipage}\hfill%
  \begin{minipage}[b]{0.49\linewidth}
  \scriptsize
  \setlength{\tabcolsep}{2pt}
  \renewcommand{\arraystretch}{1.1}

  \begin{tabular*}{\linewidth}[b]
      {@{\extracolsep{\fill}}lcc@{}}
  \toprule
  \textbf{Method}
  & \textbf{Rot. Err.} $\downarrow$
  & \textbf{Trans. Acc.} $\uparrow$ \\
  \midrule
  w/o structured semantic control
  & 0.478 & 88.9 \\
  Ours (Bidirectional)
  & \textbf{0.168} & \textbf{98.3} \\
  \bottomrule
  \end{tabular*}
  \end{minipage}

  % 两组表之间的距离
  \par\nointerlineskip
  % \vspace{5pt}
  % ========================================================
    % 第二组：标题顶部对齐，表格底部对齐
    % ========================================================
    \begingroup

    % 左列：先保存标题和表格，测量自然高度
    \setbox0=\vbox{%
      \hbox{%
        \begin{minipage}[t]{0.40\linewidth}
          \vspace{0pt}
          \caption{Quantitative comparison on memory.
          For fair comparisons, both variants utilize a bidirectional backbone and train under 96 latents.}
          \label{tab:ablation_memory}
        \end{minipage}%
      }%
      \nointerlineskip
      \vskip 4pt plus 1fill
      \hbox{%
        \begin{minipage}[b]{0.40\linewidth}
          \scriptsize
          \setlength{\tabcolsep}{2pt}
          \renewcommand{\arraystretch}{1.1}

          \begin{tabular*}{\linewidth}[b]
            {@{\extracolsep{\fill}}lcccc@{}}
            \toprule
            \textbf{Method}
            & \textbf{PSNR} $\uparrow$
            & \textbf{SSIM} $\uparrow$
            & \textbf{LPIPS} $\downarrow$
            & \textbf{MEt3R} $\downarrow$ \\
            \midrule
            Full-context
            & 18.47 & \textbf{0.688} & \textbf{0.406} & 0.133 \\
            Ours
            & \textbf{18.80} & 0.653 & 0.414 & \textbf{0.128} \\
            \bottomrule
          \end{tabular*}
        \end{minipage}%
      }%
      \kern0pt
    }%

    % 右列：先保存标题和表格，测量自然高度
    \setbox2=\vbox{%
      \hbox{%
        \begin{minipage}[t]{0.58\linewidth}
          \vspace{0pt}
          \caption{Quantitative comparison on distillation.}
          \label{tab:ablation_distillation}
        \end{minipage}%
      }%
      \nointerlineskip
      \vskip 4pt plus 1fill
      \hbox{%
        \begin{minipage}[b]{0.58\linewidth}
          \scriptsize
          \setlength{\tabcolsep}{2pt}
          \renewcommand{\arraystretch}{1.1}

          \begin{tabular*}{\linewidth}[b]
            {@{\extracolsep{\fill}}lcccccc@{}}
            \toprule
            & \multicolumn{6}{c}{\textbf{WBench}} \\
            \cmidrule(l){2-7}
            \textbf{Method}
            & \textbf{Avg.} $\uparrow$
            & \textbf{Qua.} $\uparrow$
            & \textbf{Set.} $\uparrow$
            & \textbf{Int.} $\uparrow$
            & \textbf{Con.} $\uparrow$
            & \textbf{Phy.} $\uparrow$ \\
            \midrule
            w/o init \& full-rollout
            & 73.9 & 74.4 & 64.6 & 81.9 & 82.6 & 65.8 \\
            w/o init
            & 75.2 & 76.2 & 68.0 & 82.6 & 83.3 & 65.9 \\
            Full-context Teacher
            & 82.3 & 81.6 & 79.5 & 88.1 & 89.4 & 72.9 \\
            Ours
            & \textbf{83.1}
            & \textbf{81.8}
            & \textbf{81.5}
            & \textbf{88.3}
            & \textbf{90.0}
            & \textbf{74.0} \\
            \bottomrule
          \end{tabular*}
        \end{minipage}%
      }%
      \kern0pt
    }%

    % 取左右两列自然高度的最大值
    \dimen0=\dimexpr\ht0+\dp0\relax
    \dimen2=\dimexpr\ht2+\dp2\relax
    \ifdim\dimen2>\dimen0
      \dimen0=\dimen2
    \fi

    % 按相同高度重新排版，差额由标题与表格之间的弹性间距吸收
    \noindent
    \vbox to \dimen0{\unvbox0}%
    \hfill
    \vbox to \dimen0{\unvbox2}%
    \par

    \endgroup

\end{table}
  
\textbf{Memory.} As reported in Tab.~\ref{tab:ablation_memory}, we evaluate long-horizon geometric consistency on revisit trajectories, comparing our compressed memory against the full-context baseline, where both variants are trained on 96 latents. While retaining uncompressed, full-resolution history endows high expressive capacity, training this model incurs prohibitive computational overhead and training time. In contrast, our compressed memory mechanism models long-horizon information in a resource-efficient manner, achieving highly competitive geometric consistency while drastically alleviating the training burden.

\textbf{Distillation.} To validate our distillation, we quantitatively benchmark different variants on WBench, as summarized in Tab.~\ref{tab:ablation_distillation}. Without few-step initialization and full-rollout replay, the student model suffers from severe mode collapse, leading to degraded visual quality (as shown in Quality dimension). While incorporating full-rollout replay partially mitigates training divergence, it remains visibly inferior to Stable Forcing across the metrics. Although distilling with the full-context teacher obtains competitive performance, scaling it to longer horizons suffers from prohibitive computational overhead, hindering efficient training. In contrast, our method achieves efficient distillation while maintaining generation quality, highlighting the effectiveness of Stable Forcing.

\end{document}